\documentclass[11pt]{article}
\usepackage{acl}
\usepackage{times}
\usepackage{latexsym}
\usepackage[T1]{fontenc}
\usepackage[utf8]{inputenc}
\usepackage{microtype}
\usepackage{inconsolata}
\usepackage{graphicx}
\usepackage{amsmath,amssymb}
\usepackage{xcolor}
\usepackage{booktabs}
\usepackage{multirow}
\usepackage{array}
\newcommand{\est}[1]{#1}
\title{MedRSI: Recursive Self-Improvement for Medical Agents via Clinically Aligned Self-Evolution}
\author{Junde Wu$^{1}$ \quad Jiayuan Zhu$^{1}$ \quad Minghao Hu$^{1}$ \quad
  Fenglin Liu$^{1}$ \quad Jiazhen Pan$^{2}$\\
  $^{1}$University of Oxford \qquad $^{2}$Stanford University}
\hypersetup{
  pdftitle={MedRSI: Recursive Self-Improvement for Medical Agents via Clinically Aligned Self-Evolution},
  pdfauthor={Junde Wu, Jiayuan Zhu, Minghao Hu, Fenglin Liu, Jiazhen Pan}
}

\begin{document}
\maketitle

\begin{abstract}
Medical agents increasingly combine general reasoning models with specialized clinical tools, yet their capabilities remain largely fixed by what clinicians and engineers design before deployment. Recursive self-improvement (RSI) offers a different paradigm in which agents learn from their own failures and autonomously expand their capabilities, but directly applying RSI to medicine creates fundamental safety challenges. We introduce \textbf{MedRSI}, the first recursive self-improvement framework for medicine, which continuously transforms diagnostic failures into new clinical capabilities through tool composition and task-specific model training. Drawing from how clinical practice prioritizes patient risk and validates new interventions before adoption, we introduce two unique mechanisms in \textbf{MedRSI}. \emph{Clinical-cost-aware failure prioritization} directs the agent's improvement effort toward failures according to their potential clinical consequences rather than their frequency alone. \emph{Fast discovery with slow registration} separates rapid capability invention from conservative adoption, allowing new tools to become part of the persistent agent only after demonstrating sustained benefit across subsequent patient cohorts. Across public glaucoma and heart disease benchmarks and two private clinical tasks, MedRSI progressively develops segmentation, measurement, prediction, multimodal reasoning, and generative capabilities, surpasses manually engineered medical agents, and autonomously discovers solutions to clinical problems that its original designers did not anticipate. MedRSI demonstrates a path toward medical agents that are not limited to capabilities specified before deployment, but can continuously construct and accumulate new capabilities from diagnostic experience while grounding what they improve and what they retain in principles derived from clinical practice. Our code is available at \url{https://github.com/ImprintLab/MedRSI}.
\end{abstract}

\section{Introduction}\label{sec-introduction}

Medical agents have begun to change how artificial intelligence participates in diagnosis. Multimodal foundation models can answer questions about medical images and records \cite{llavamed,medflamingo,medpalmm,biomedclip}, but they often lack the fine-grained perception and quantitative analysis that diagnosis requires \cite{medagentpro}. In agentic systems, a reasoning model organizes a clinical investigation, retrieves guideline knowledge and delegates specialized subtasks to expert tools such as segmentation networks, measurement code and visual question answering models \cite{mmedagent,medrax,medagentpro,macro,evomedagent}, or several models collaborate through role-play, debate and adaptive team formation \cite{medagents,mdagents,kg4diagnosis,agenthospital}. For example, MedAgent-Pro illustrates this design for glaucoma and heart disease diagnosis, where guideline-derived plans invoke optic disc and cup segmentation with the Medical SAM Adapter \cite{msa}, cup-to-disc ratio calculation and ventricular quantification, and where the resulting agent outperforms single multimodal foundation models by wide margins\cite{medagentpro}. The intelligence of such an agent is the sum of a general reasoning model and a fixed set of human-designed tools. Every tool was conceived by a clinician who recognized a diagnostic need, trained by an engineer who assembled data and code, and integrated by a developer who wrote its calling specification. Once deployed, the agent accumulates thousands of diagnostic encounters and failures, and it has no mechanism to transform this experience into new clinical capabilities.

Recursive self-improvement (RSI) has recently emerged as a fundamentally different paradigm for building increasingly capable agents. Language agents already refine their outputs and plans from feedback \cite{react,reflexion,selfrefine}, learn to call external tools \cite{toolformer}, and can write new tools or reusable skills for themselves \cite{latm,creator,voyager}. Rather than relying on human engineers to continually diagnose failures, design new tools, and manually refine increasingly complex agent harnesses, RSI shifts this process to the agent itself. The agent repeatedly evaluates its own failures, identifies limitations in its current capabilities or execution environment, proposes modifications, and retains those that improve future performance \cite{stop,godelagent,dgm}. Recent frameworks have shown that such autonomous evolution can discover agent designs, harness configurations and algorithms that even outperform carefully engineered human-designed systems \cite{adas,dgm,alphaevolve}. This not only reduces the substantial human effort required to develop and maintain increasingly complex agent harnesses, but also suggests a more fundamental possibility that an agent can continually reshape its own interaction mechanisms as it encounters new tasks and failures. RSI therefore opens a path toward persistent self-evolution which increasingly general capabilities emerge through an open-ended process of autonomous self-improvement.

\begin{figure*}[!t]
\centering
\includegraphics[width=\textwidth]{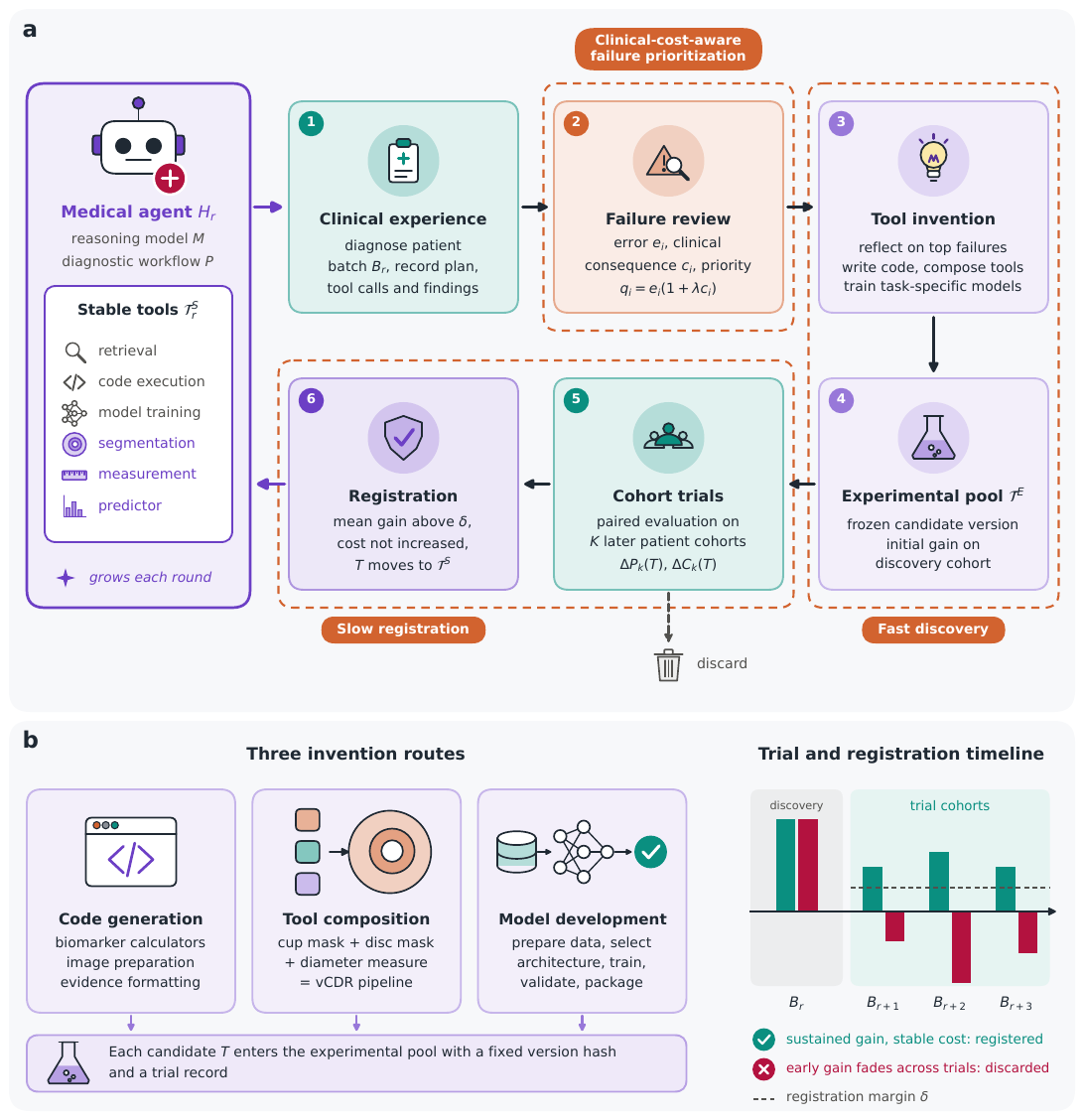}
\caption{\textbf{Recursive self-improvement of a medical agent in MedRSI.} \textbf{a}, The medical agent $H_r$ combines a reasoning model, a diagnostic workflow and a stable tool registry that grows across rounds. In each round the agent diagnoses a patient batch, reviews its failures with clinical consequence weighting, invents candidate tools, places them in an experimental pool and evaluates them on later cohorts before registration. \textbf{b}, Candidate tools arise from code generation, tool composition and model development. Each candidate is frozen with a version hash and tracked across a discovery cohort and $K$ trial cohorts. A candidate with sustained gain and stable clinical cost is registered, and a candidate whose early gain fades is discarded.}\label{fig-framework}
\end{figure*}

However, directly extending RSI to risk-sensitive domains such as medicine demands substantially more careful design. As an AI system continuously modifies and improves itself toward a predefined objective, the resulting behaviors may not remain within the boundaries originally anticipated by its human designers. An update that improves measured performance can simultaneously introduce unexpected failure modes, unsafe strategies, or behaviors that exploit imperfections in the objective itself \cite{concreteproblems}. Such risks become particularly consequential in medicine, where an autonomously evolved decision rule, tool, or workflow can directly affect clinical decisions and patient outcomes. A failure that appears as a small performance regression in a general-purpose benchmark may instead translate into a missed diagnosis, an inappropriate intervention, or other forms of patient harm. More concerningly, because self-improving agents can repeatedly retain and propagate their own modifications, an unsafe change may become embedded in subsequent generations of the system and potentially affect patients at scale. RSI in medicine therefore introduces a fundamental tension between allowing agents sufficient autonomy to discover capabilities beyond human-designed harnesses and ensuring that this open-ended evolution remains aligned with clinical safety.

We introduce \textbf{MedRSI}, the first recursive self-improvement framework for medicine (Fig.~\ref{fig-framework}). MedRSI continuously transforms diagnostic failures into new medical capabilities by identifying recurrent failure patterns, reflecting on the capabilities needed to address them, and creating new tools through tool composition and task-specific model training. We introduce two unique designs motivated by practical clinical considerations. First, diagnostic failures are not equally consequential. A minor diagnostic disagreement and a missed life-threatening disease can carry radically different risks to the patient. We therefore introduce \emph{clinical-cost-aware failure prioritization}, which directs the agent's self-improvement toward failures according to their clinical consequences rather than their frequency alone. Second, a newly discovered capability should not immediately become a permanent part of a medical agent. Much like a new medical device or intervention requires clinical evaluation before deployment, a tool that appears beneficial on a limited set of cases may prove unstable or even harmful when applied to new patients or combined with other tools. We therefore introduce \emph{fast discovery with slow registration}, which allows the agent to rapidly create and experiment with candidate tools while admitting them into its persistent toolkit only after their benefits remain consistent across subsequent patient cohorts. Together, these designs enable continuous autonomous capability expansion while ensuring that both what the agent improves and what it permanently retains are grounded in clinical consequence and empirical evidence.

This work makes three contributions. First, we introduce MedRSI, a medical recursive self-improvement framework that enables a diagnostic agent to autonomously invent, evaluate and accumulate new task-specific clinical tools and AI models from its own diagnostic failures. Second, we introduce clinical-cost-aware improvement, which considers both whether a prediction is wrong and how consequential the error would be in practice. Third, we introduce fast discovery and slow consolidation, which admits new capabilities into the medical agent only after sustained improvement across subsequent patient cohorts. Across public glaucoma and heart disease benchmarks and two private multimodal clinical tasks, MedRSI grows from a general agent into a system that surpasses a manually engineered medical agent, rediscovers the capability classes that experts previously built by hand, invents new multimodal and generative models for tasks it has never encountered, and remains stable across long improvement horizons.

\section{Method}\label{sec-methods}

\subsection{Overall Workflow of MedRSI}

MedRSI turns diagnostic experience into persistent new capabilities through a recurring five-stage cycle: \emph{diagnosis}, \emph{prioritization}, \emph{reflection}, \emph{capability development}, and \emph{trial and registration}. The stable medical agent first diagnoses a batch of patients using its current toolkit. An evaluator then identifies failures and prioritizes them according to their clinical consequences. Related failures are grouped and reflected upon to identify the capability that is missing or unreliable. A tool builder translates this specification into executable candidates through code generation, tool composition, or task-specific model training. Candidate capabilities are tested on subsequent patient cohorts and become part of the stable agent only when their benefit persists. The next round therefore begins with an agent whose capabilities reflect the improvements validated in previous rounds. Repeating this cycle allows diagnostic experience to progressively reshape what the agent can observe, measure, predict, and integrate.

At improvement round $r$, the stable medical agent is represented by
\begin{equation}
H_r=(M,P,\mathcal{T}^{S}_r),
\label{eq-agent-state}
\end{equation}
where $M$ is the multimodal reasoning model, $P$ is the general execution workflow, and $\mathcal{T}^{S}_r$ is the registry of stable tools. The parameters of $M$ and the execution rules of $P$ remain fixed throughout an improvement trajectory. Self-improvement instead operates on $\mathcal{T}^{S}_r$: registered tool descriptions, executable functions, and trained checkpoints determine the clinical operations available to the agent. A successful improvement therefore changes the evidence that the agent can obtain and use when diagnosing subsequent patients.

Concretely, each round begins with an experience batch $B_r={(x_i,y_i)}_{i=1}^{N_r}$. The stable agent diagnoses each case, producing
\begin{equation}
(\hat y_i,\tau_i)=H_r(x_i),
\end{equation}
where $\tau_i$ is the execution record containing the tool versions and calls, returned measurements, observable evidence summaries, and final prediction. Reference labels are withheld during diagnostic execution and revealed only afterward to the evaluator, ensuring that each diagnosis is committed before its reference becomes available. The resulting failures then pass through the improvement cycle: clinically consequential cases are prioritized, related failures are clustered, reflection converts each cluster into a missing-capability specification, and the builder produces candidate tools $\mathcal{C}_r$. Candidates that demonstrate an initial benefit enter an experimental pool and are evaluated on subsequent cohorts. Once a candidate demonstrates sustained benefit, it is registered into the stable toolkit and becomes available to future rounds. Thus, information flows from

\[
\begin{gathered}
\text{experience}\rightarrow\text{failure}\rightarrow\text{capability gap}\\
\rightarrow\text{candidate tool}\rightarrow\text{validated capability},
\end{gathered}
\]

with each validated capability changing the starting point of the next improvement cycle.

\paragraph{Failure evaluation and prioritization.}
After diagnosis, the evaluator determines both whether each prediction is correct and the clinical consequence of an error. The priority mechanism described in Section~\ref{sec:priority} uses these signals to allocate a bounded reflection budget. The evolver then groups related high-priority failures according to clinical task, available modalities, failed intermediate operations, and tool-use patterns. Several missed glaucoma cases, for example, may share an uncertain optic-cup boundary, whereas another group may contain reliable individual measurements that nevertheless lead to inconsistent diagnostic synthesis. The former suggests a missing perceptual capability and the latter a missing integration capability. Each resulting cluster contains representative cases, their reference labels, and the relevant portions of their execution records.

\paragraph{Reflection and capability specification.}
Reflection converts each failure cluster into a structured development specification. Rather than proposing a tool directly from an individual mistake, it identifies the clinical observation or operation that is systematically missing or unreliable, the evidence supporting this interpretation, the function that a candidate capability should provide, its required inputs, and the endpoint on which its benefit should be evaluated. The specification additionally records available supervision, compatible stable tools, and feasible development routes under the remaining budget. Reflection therefore treats the inferred capability gap as a testable hypothesis. A cluster suggesting unreliable optic-cup delineation, for example, can motivate a segmentation capability whose value is subsequently evaluated both through its anatomical output and through its effect on downstream diagnosis.

\paragraph{Capability development.}
The tool builder realizes each specification through three complementary routes: \emph{code generation}, \emph{tool composition}, and \emph{model development}. Code generation implements deterministic operations such as image preparation, quantitative measurement, calculation, and structured evidence handling, together with explicit contracts for input types, dimensions, units, missing values, and returned intermediate evidence. Tool composition combines compatible existing capabilities into higher-level clinical procedures. For example, cup and disc segmentation can be composed with diameter measurement to obtain
\begin{equation}
\mathrm{vCDR}=\frac{d_{\mathrm{cup}}}{d_{\mathrm{disc}}},
\label{eq-vcdr}
\end{equation}
with the resulting tool returning the ratio, component contours, and a quality indicator while retaining the exact dependency versions.

Model development is used when the missing capability must be learned from labelled examples. The builder constructs a permitted training manifest, selects a generic architecture, defines preprocessing and augmentation, trains the model, and selects a checkpoint on an internal development partition. Depending on the specification, this can produce segmentation networks, clinical predictors, or multimodal fusion models. Architecture choices, losses, optimization settings, stopping criteria, training data, and selected checkpoints are retained as part of the candidate record. Model fitting and checkpoint selection use only the development resources assigned to the builder; discovery and registration evaluations remain outside this fitting process.

The same mechanism can construct multi-stage capabilities when resolving one limitation requires creating an intermediate resource. For example, when labelled examples in a target imaging domain are scarce, the builder may develop a generative model, validate the anatomical consistency of its outputs, and use the resulting synthetic data to train a downstream predictor. The generator, generated-data manifest, and predictor are then retained as a versioned dependency chain, while final evaluation is performed on real held-out examinations using the downstream clinical endpoint. MedRSI can therefore improve not only by inventing a new diagnostic tool directly, but also by constructing capabilities that enable the development of subsequent capabilities.

\paragraph{Candidate packaging and experimentation.}
Every developed candidate is packaged as an inspectable artifact containing executable code, model weights where applicable, input and output schemas, a clinical calling description, dependencies, development records, and a version hash. Before evaluation, the builder performs functional checks covering execution, output structure, units, missing inputs, and unavailable measurements. The candidates generated during round $r$ form
\begin{equation}
\mathcal{C}_r=\{T_{r,1},\ldots,T_{r,m_r}\},
\end{equation}
where $m_r$ is bounded by the development budget.

Candidates that demonstrate an initial discovery benefit enter the experimental pool rather than immediately modifying the stable agent. Their trial records accumulate on subsequent patient cohorts while diagnosis continues using the stable toolkit. Once the required trial horizon is complete, accepted candidates form $\mathcal{A}_r$, and the persistent registry is updated as
\begin{equation}
\mathcal{T}^{S}_{r+1}
=
\mathcal{T}^{S}_r\cup\mathcal{A}_r,
\qquad
H_{r+1}
=
(M,P,\mathcal{T}^{S}_{r+1}).
\label{eq-registry-update}
\end{equation}
For an accepted replacement, the registry points to the new version while retaining its predecessor in the development history. If no candidate qualifies, the stable toolkit remains unchanged. The next experience batch is then processed by the resulting agent, and failures are reassessed under its expanded capability set. A capability gap that dominated earlier rounds can therefore disappear once an effective tool has been acquired, allowing subsequent improvement to move toward unresolved limitations.

Two clinically motivated mechanisms govern this cycle at different decision points. \emph{Clinical-cost-aware failure prioritization} determines which diagnostic experiences receive the agent's limited improvement effort and therefore shapes \emph{what the agent attempts to improve}. \emph{Fast discovery with slow registration} separates exploratory capability development from changes to the persistent agent and therefore determines \emph{what the agent is allowed to retain}. We describe these two mechanisms in detail next.

\subsection{Clinical-Cost-Aware Failure Prioritization}\label{sec:priority}

Not every diagnostic failure should contribute equally to self-improvement. A common borderline disagreement may have little effect on patient management, whereas a less frequent error can delay treatment, miss progressive disease, or incorrectly discharge a high-risk patient. If RSI allocates its limited development budget according only to error frequency, it can become increasingly effective at correcting common mistakes while repeatedly overlooking failures with greater potential patient impact. We therefore make clinical consequence an explicit signal for deciding which experiences should drive capability development.

The evaluator separates two questions: \emph{was the diagnosis wrong?} and, if so, \emph{how consequential could that error be?} Correctness is determined against the task reference, while consequence is assessed from the change in clinical action implied by the discrepancy under a prespecified task-specific rubric. This separation preserves the conventional diagnostic endpoint while providing the evolver with an additional signal describing which failures matter most clinically.

For classification, we define the error indicator as
\begin{equation}
e_i=\mathbf{1}(\hat y_i\neq y_i).
\end{equation}
For each error, the clinical judge receives the available patient observations, reference diagnosis, committed prediction, and corresponding execution record. A task-specific rubric specifies the clinical actions associated with diagnostic outputs, such as discharge, routine reassessment, further investigation, or specialist review. The judge evaluates how the erroneous prediction would alter this pathway and assigns an ordinal consequence score
\begin{equation}
s_i\in{0,1,2,3,4},
\end{equation}
corresponding to \emph{negligible}, \emph{minor}, \emph{moderate}, \emph{serious}, and \emph{critical} potential consequences. The assessment is evidence-linked: together with the score, the judge records the clinical observations supporting it, the implied change in management, and any information required but unavailable.

The rubric is instantiated for each clinical task before self-improvement begins. For glaucoma, consequence depends on the documented disease stage, predicted category, and resulting follow-up pathway. Missing advanced glaucoma with discharge, for example, carries a different consequence from assigning a borderline eye to short-term reassessment. For heart disease, the rubric additionally considers functional impairment and the investigation or review implied by the agent's prediction. When the available evidence is insufficient to determine consequence, the case is marked unresolved and withheld from consequence-based prioritization until adjudicated.

To keep this signal fixed across the self-improvement trajectory, we use a frozen judge prompt, model version, rubric, and output schema. Clinicians independently score a calibration set, adjudicate disagreements, and review discrepancies between the judge and adjudicated references; ordinal agreement is summarized using weighted $\kappa$. The resulting evaluator configuration is then frozen across rounds and ablations, and agent identity is hidden during paired candidate evaluation.

We normalize the consequence score as
\begin{equation}
c_i=\frac{s_i}{4},
\end{equation}
and assign each experience the failure priority
\begin{equation}
q_i=e_i(1+\lambda c_i),
\label{eq-priority}
\end{equation}
where $\lambda\geq0$ controls how strongly clinical consequence influences self-improvement. Every diagnostic error retains a base priority of one, while clinically consequential errors receive additional weight proportional to their assessed potential harm. Thus, for an erroneous prediction, $q_i\in[1,1+\lambda]$; correct predictions receive zero failure priority but remain available for evaluating candidate performance. Setting $\lambda=0$ recovers conventional uniform error prioritization and defines the corresponding ablation.

Consider two glaucoma failures under $\lambda=2$. An early glaucoma case classified as suspect with prompt reassessment may receive a minor consequence score, $c_i=0.25$, giving $q_i=1.5$. An eye with documented advanced damage classified as normal and discharged may receive a serious score, $c_i=0.75$, giving $q_i=2.5$. Both predictions are incorrect, but the latter receives substantially greater improvement priority because the resulting clinical pathway carries greater potential consequence. The precise grades are determined by the prespecified rubric and documented clinical context rather than by the diagnostic label alone.

The resulting priorities determine which experiences enter the bounded reflection buffer. Failures are first grouped by their inferred capability gaps, and cases within each cluster are ordered by $q_i$. The selector takes the highest-priority representative from each cluster before returning to lower-ranked cases, ensuring that high-consequence failures receive early attention without allowing a single recurrent failure mode to consume the entire development budget. The evolver receives both the consequence score and its evidence-linked justification, connecting the capability it proposes to the clinical failure that motivated its development.

Priorities are recomputed after every experience batch. Once an acquired capability resolves a failure, correctly diagnosed instances of that pattern no longer enter error-driven reflection; unresolved failures continue to compete for development effort according to their clinical consequences. Clinical-cost-aware prioritization therefore continuously redirects the finite self-improvement budget toward the most consequential limitations of the current agent rather than toward a fixed set of errors.

We use the same consequence assessment to quantify whether self-improvement reduces clinically important errors at the cohort level. For an evaluation cohort $V$ containing $n_V$ patients, we define
\begin{equation}
C(H,V)
=
\frac{1}{n_V}
\sum_{i\in V}
e_i(H)c_i(H),
\label{eq-clinical-cost}
\end{equation}
and report $100\,C(H,V)$ as clinical cost per 100 patients. We additionally report the frequency of serious and critical errors, defined as the proportion of evaluated patients for whom $e_i=1$ and $s_i\geq3$. Because consequence is reassessed for the prediction produced by each compared agent under the same reference and rubric, an improvement can reduce clinical cost either by correcting an error entirely or by replacing it with a less consequential residual error. This makes clinical cost a shared quantity connecting which failures MedRSI chooses to improve with whether the resulting capability actually reduces clinically meaningful mistakes.

For continuous ejection-fraction prediction, the same formulation is applied after defining correctness through a prespecified clinical tolerance. Given absolute error
\begin{equation}
a_i=|\hat y_i-y_i|,
\end{equation}
we set
\begin{equation}
e_i=\mathbf{1}(a_i>\epsilon),
\end{equation}
with $\epsilon=\est{5}$ ejection-fraction percentage points in our experiments. The task-specific rubric then assigns consequence according to error magnitude, direction, and documented clinical context, with prespecified boundaries at \est{5}, \est{10}, \est{15}, and \est{20} percentage points. Normalized mean absolute error remains the primary continuous prediction endpoint, while the consequence score captures the clinical significance of the remaining discrepancy.

\subsection{Fast Discovery and Slow Tool Registration}\label{sec:registration}

A capability that appears useful when first discovered should not immediately become part of a persistent medical agent. Candidate tools are often developed from a small set of difficult cases, and an improvement on those cases may reflect cohort-specific imaging characteristics, disease presentations, or interactions with the agent's current toolkit rather than a generalizable clinical benefit. In medicine, this distinction motivates staged evaluation before a new intervention or device becomes established practice. MedRSI adopts the same principle for autonomously developed capabilities: discovery provides evidence that a candidate is worth testing, but persistent adoption requires evidence that its benefit survives subsequent patient cohorts.

This distinction is particularly important for recursive self-improvement. Once a tool enters the stable registry, its effect extends beyond the patients on which it is immediately used. The planner can invoke it in future diagnoses, its outputs become part of subsequent execution records, and those records in turn determine which failures the agent reflects upon and which capabilities it attempts to develop next. An unstable tool can therefore alter not only current predictions but also the future direction of self-improvement. MedRSI separates rapid capability invention from conservative capability adoption through \emph{fast discovery with slow registration}: candidates can be generated and explored aggressively, while the persistent agent changes only after their benefit has been reproduced across subsequent patient cohorts.

MedRSI maintains two tool sets: an experimental pool $\mathcal{T}^{E}_r$ containing capabilities under evaluation and a stable registry $\mathcal{T}^{S}_r$ containing capabilities available to the deployed improvement trajectory. The stable agent uses only $\mathcal{T}^{S}_r$ when collecting new diagnostic experience. Experimental tools are exposed only to evaluator copies of the agent for controlled comparison. This separation allows the search process to remain exploratory without allowing every promising discovery to immediately influence future diagnoses and self-improvement.

After passing the functional checks described above, a candidate $T$ undergoes a discovery evaluation. The evaluator constructs a matched copy of the current stable agent in which the candidate's calling description and executable entry point are additionally available to the planner. The stable and candidate-augmented agents process the same discovery cases under matched inference settings. We denote the resulting change in the prespecified task endpoint by $\Delta P_{\mathrm{disc}}(T)$. A candidate satisfying
\begin{equation}
\Delta P_{\mathrm{disc}}(T)>0
\end{equation}
enters the experimental pool. This deliberately permissive criterion treats discovery as a screening stage: its purpose is to retain potentially useful capabilities for further evaluation rather than to establish sufficient evidence for permanent adoption.

Admission to the experimental pool freezes the complete candidate package, including its executable code, model weights, preprocessing, calling specification, and dependency versions. Any subsequent modification creates a new candidate version with a new evaluation history. The reference stable-registry snapshot is also recorded, ensuring that all evidence collected for a candidate refers to the exact artifact and agent context under evaluation.

A frozen candidate is then evaluated prospectively with respect to the improvement trajectory on $K$ subsequent patient cohorts that were not used for candidate fitting or discovery. Let $H^{\mathrm{ref}}_T$ denote the stable-agent snapshot associated with candidate $T$, and let $V_{T,1},\ldots,V_{T,K}$ denote its trial cohorts. Within each cohort, the evaluator compares the reference agent with the same agent augmented by the candidate,

$$
H^{\mathrm{ref}}_T
\qquad\text{and}\qquad
H^{\mathrm{ref}}_T\oplus T,
$$

where $\oplus T$ makes the candidate available for ordinary planner selection rather than forcing its use. Candidate calls, execution failures, and cases in which the planner chooses not to invoke the candidate all remain part of the evaluation, so the measured effect reflects the practical contribution of adding the capability to the complete diagnostic workflow.

For trial cohort $k$, the paired change in diagnostic performance is
\begin{equation}
\begin{aligned}
\Delta P_k(T)
={}&P(H^{\mathrm{ref}}_T\oplus T,V_{T,k})\\
&-P(H^{\mathrm{ref}}_T,V_{T,k}),
\end{aligned}
\label{eq-trial-benefit}
\end{equation}
where $P$ is balanced accuracy for classification and negative normalized mean absolute error for continuous ejection-fraction prediction, such that larger values always indicate better performance. In parallel, we measure the change in clinical cost,
\begin{equation}
\begin{aligned}
\Delta C_k(T)
={}&C(H^{\mathrm{ref}}_T\oplus T,V_{T,k})\\
&-C(H^{\mathrm{ref}}_T,V_{T,k}).
\end{aligned}
\label{eq-trial-cost}
\end{equation}
The same clinical judge and rubric described in Section~\ref{sec:priority} are applied to both configurations. Because patient inputs, inference settings, and stochastic seeds are matched, these paired evaluations isolate the effect of making the candidate capability available to the agent.

The registration horizon $K$ counts evaluable cohorts rather than improvement rounds. Candidate trials therefore proceed alongside continued self-improvement: a round can generate new candidates while advancing the trial histories of candidates discovered earlier. Trial cohorts contain disjoint patients within each candidate evaluation and satisfy prespecified eligibility, minimum-size, and class-coverage requirements. Candidates remain pending when an eligible cohort is unavailable rather than being registered with incomplete evidence.

After completing the trial horizon, a candidate is eligible for registration only if it improves diagnostic performance while introducing no increase in clinical cost:
\begin{equation}
\begin{aligned}
\overline{\Delta P}(T)
&=
\frac{1}{K}\sum_{k=1}^{K}\Delta P_k(T)>\delta,\\
\overline{\Delta C}(T)
&=
\frac{1}{K}\sum_{k=1}^{K}\Delta C_k(T)\leq0,
\end{aligned}
\label{eq-registration}
\end{equation}
where $\delta$ is a prespecified minimum improvement margin. The first condition requires the candidate to demonstrate sustained diagnostic benefit beyond its original discovery cases; the second prevents an apparent aggregate performance gain from being purchased through more clinically consequential residual errors.

Our default configuration uses $K=3$ subsequent trial cohorts. For classification, $\delta=0.005$ corresponds to a balanced-accuracy improvement of 0.5 percentage points. For example, trial gains of 1.2, 0.6, and 0.9 percentage points yield a mean improvement of 0.9 points and satisfy the performance criterion, provided clinical cost does not increase. In contrast, a candidate with a large initial discovery gain followed by trial changes of $+3$, $-1$, and $-2$ points has zero mean subsequent benefit and is rejected. Discovery therefore asks whether a capability is promising enough to investigate, whereas registration asks whether its benefit persists beyond the experience that motivated its creation.

A qualifying candidate moves from $\mathcal{T}^{E}_r$ into the stable registry together with its calling specification, dependencies, development record, and complete trial history. Rejected candidates are archived with their evaluation outcomes, which can inform subsequent reflection and development, but their outputs never enter the stable diagnostic trajectories. The stable agent therefore continues operating with the last validated registry while experimentation proceeds in parallel.

Because MedRSI evolves cumulatively, a candidate's contribution depends on the capabilities already present when it is introduced. A new segmentation model can alter measurements consumed by downstream predictors, while a new predictor can change how the planner uses existing tools. We therefore commit accepted capabilities serially. Before registration, the candidate's reference-registry hash is compared with the current stable registry. If the registry has changed during its trial period, the candidate is reevaluated in the updated context before being admitted. Updates to shared components are packaged together with affected dependent functions and evaluated as a single versioned change.

This mechanism gives slow registration a role beyond conventional model validation. It protects the future self-improvement process itself. A candidate that fails to generalize is prevented from generating the diagnostic trajectories on which later reflection and invention would depend, while a validated capability becomes a stable foundation from which further capabilities can be constructed. Fast discovery therefore preserves the exploratory character of RSI, whereas slow registration controls which discoveries are allowed to influence its future. Their separation enables MedRSI to expand its capabilities aggressively without allowing transient improvements to recursively accumulate into persistent degradation.

\section{Experiments}

\subsection{Datasets and patient separation}

The public tasks follow the clinical endpoints examined by MedAgent-Pro\cite{medagentpro}. REFUGE2 provides fundus photographs with glaucoma labels and optic disc and cup annotations\cite{refuge2}, and the MedAgent-Pro evaluation resource contains 1,200 images\cite{medagentprodetails}. MITEA contains 536 three-dimensional echocardiographic images from 143 subjects with left ventricular annotations derived from subject-specific cardiac magnetic resonance, and its seven cardiac categories are reduced to healthy and heart disease for binary evaluation\cite{mitea,medagentpro}. The multimodal glaucoma cohort was assembled from \est{two} tertiary eye centres between \est{2018 and 2024} and contains \est{1,898} eyes from \est{1,236} patients. Eyes were eligible when at least two of three examinations were acquired on the same day, namely a fundus photograph, spectral-domain optical coherence tomography of the peripapillary retinal nerve fibre layer and a reliable 24-2 visual field, in line with guideline recommendations for structural and functional assessment\cite{niceglaucoma}. All three modalities are available for \est{93\%} of eyes, and the remaining \est{7\%} lack one modality, most often an unreliable visual field, which is represented explicitly through modality masks. The reference diagnosis was adjudicated by \est{two} glaucoma specialists with a third resolving disagreements. The contrast echocardiography cohort contains \est{318} contrast-enhanced and \est{2,940} noncontrast transthoracic studies from \est{1,874} patients acquired between \est{2016 and 2024}, and the reference ejection fraction is derived from cardiac magnetic resonance performed within \est{30} days of the echocardiogram. \est{Forty-two} patients have a same-day paired contrast and noncontrast study, and both studies of these patients are withheld from all training, discovery, trial and test partitions and used only for evaluating the translation model.

Patient identity is the unit of separation. All eyes, visits, phases, volumes and derived images from a patient remain within one partition. Development data are divided into model training, clinical experience batches, discovery validation and registration trial cohorts, and a locked final test partition is held separately. For REFUGE2 the split is \est{600} training, \est{200} discovery, \est{200} trial and \est{200} test images, and for MITEA it is \est{86} training, \est{19} discovery, \est{19} trial and \est{19} subjects for test. For multimodal glaucoma, \est{486} eyes from \est{312} patients form the test partition, and the remaining \est{1,412} development eyes are divided into \est{892} eyes for model training and experience batches, \est{130} for discovery and three trial cohorts of \est{130} eyes each. For contrast echocardiography, the \est{318} contrast studies comprise \est{96} test studies, the \est{42} paired translation-evaluation studies, \est{120} training studies, \est{15} discovery studies and three trial cohorts of \est{15} studies. Noncontrast studies from patients in any contrast partition other than training are excluded, leaving \est{2,716} of the \est{2,940} noncontrast studies available for training. Discovery and trial evaluation on the private tasks uses contrast or multimodal cases only. Each round draws a clinical experience batch of \est{120} glaucoma and \est{60} heart disease cases from the training partition. The builder receives training labels and completed experience records. Candidate selection uses discovery and trial results only. Test outcomes are computed after the full improvement trajectory and its checkpoints have been frozen, and the round-wise test curves are generated retrospectively from those checkpoints. Generated training images inherit the partition and patient provenance of their source examinations.

\subsection{Implementation and comparators}

The initial harness is \est{OpenHands v0.20} with its executable-code agent, run in its containerized workspace with network access limited to the permitted retrieval sources\cite{openhands}. All agents use \est{GPT-4o (version 2024-08-06)} as the reasoning model \cite{gpt4o}, which is the core model of the source MedAgent-Pro \cite{medagentpro}, with temperature \est{0} for diagnosis and \est{0.7} for reflection and tool building. The clinical judge uses \est{a separately prompted instance of the same model} with the fixed rubric. Baseline agents and MedRSI share identical inference settings, permitted training resources and patient partitions. The public-task protocol contains 20 improvement rounds, extended to 30 for the long-horizon ablation, and each round permits up to \est{six} candidate development attempts with a shared compute allowance of \est{12} accelerator hours on \est{NVIDIA A100 80 GB} devices. Model development uses \est{U-Net \cite{unet}, nnU-Net \cite{nnunet}, ResNet-50 \cite{resnet}, a three-branch fusion transformer and a CycleGAN-style translation network \cite{cyclegan}} as the architecture library, and the agent selects architectures and hyperparameters itself. The complete configuration uses $\lambda=2$, $K=3$ trial cohorts and a registration margin $\delta$ of \est{0.5} balanced accuracy points for classification and \est{0.01} normalized error units for ejection fraction prediction. All protocol settings were fixed before access to the final test set. The comparator table preserves the values reported by MedAgent-Pro \cite{medagentpro} for general multimodal models \cite{gpt4o,janus,llavamed,biomedclip,qwen25vl,internvl25}, medical agents \cite{medagents,mmedagent,mdagents}, ophthalmology models \cite{retizero,visionunite} and the REFUGE2 challenge leaderboard \cite{refuge2}. In that evaluation, agents designed for text-based questions were adapted to visual question answering with their core collaboration mechanism, and all agentic baselines used GPT-4o as their underlying model. Because published values do not provide patient-level predictions, we also reran the released MedAgent-Pro pipeline on our REFUGE2 and MITEA test partitions with the same reasoning model version, inference settings and \est{five} seeds; these reruns are used only for paired significance testing, and their balanced accuracies of \est{89.6\%} for glaucoma and \est{77.1\%} for heart disease are within \est{0.8} points of the published values. The task-specific models on the private tasks are a three-branch fusion network for multimodal glaucoma and a three-dimensional ResNet regressor \cite{resnet3d} for ejection fraction, each trained by the study team on the same labelled training cases with a development budget equal to the total budget consumed by MedRSI on that task.

\subsection{Statistical analysis and development records}

Balanced accuracy is the mean of sensitivity and specificity, and F1 is the harmonic mean of precision and recall for the disease-positive class. Normalized mean absolute error divides the mean absolute error by the training-set standard deviation of the endpoint. Repeated examinations are aggregated by a prespecified patient-level rule before analysis. Each complete improvement trajectory is repeated with \est{five} random seeds, and patient-level bootstrap resampling with \est{2,000} replicates estimates 95\% confidence intervals for performance and paired endpoint differences. Comparisons between MedRSI and its ablations, and between MedRSI and the MedAgent-Pro reruns, use paired permutation tests on patient-level predictions with \est{10,000} permutations. The final round is the prespecified primary comparison and earlier checkpoints describe the trajectory.

Tool discovery analysis records for every candidate the initiating failure cluster, generated code, training configuration, discovery endpoint, cohort trial outcomes and registration decision. The category codebook for capability coverage was written from the MedAgent-Pro description of its toolkit before any generated tool was reviewed, and two authors assigned categories independently with \est{full} agreement on the registered tools. The number of stable tools counts independently callable task-specific functions, with shared dependencies tracked separately. Segmentation quality of self-trained models is measured with the Dice coefficient against the public annotations, and generated contrast images are compared with paired real frames using the structural similarity index.

\subsection{Results}\label{sec-results}

\subsection{A general agent develops medical diagnostic capabilities through recursive self-improvement}

The first experiment asks a general agent to build itself into a medical diagnostic system. The initial agent $H_0$ is OpenHands, a public open-source platform for generalist agents, which executes its actions as code inside a sandboxed workspace\cite{openhands,codeact}. The platform supplies a multimodal reasoning model, web and document retrieval, a shell, a Python interpreter, read access to the permitted training data and the standard scientific Python stack, through which model training is available as an ordinary code action. Its capabilities are those of a general software and tool-use agent, and it carries no clinical content. The evaluation follows the MedAgent-Pro public setting, with glaucoma diagnosis on REFUGE2 fundus photographs and heart disease diagnosis on MITEA three-dimensional echocardiography\cite{medagentpro,refuge2,mitea}. We set the agent to improve for 20 rounds on the training partition, all checkpoints are frozen, and test performance is then computed retrospectively for every round.

\begin{table*}[!t]
\caption{\textbf{Diagnostic performance on the public tasks.} Balanced accuracy (bAcc), F1 and area under the receiver operating characteristic curve (AUC) are percentages. Comparator values are reported by MedAgent-Pro\cite{medagentpro}, including its evaluation of the ophthalmology models and the REFUGE2 challenge leaderboard. MedRSI values are means over \est{five} independent improvement trajectories, and their confidence intervals are given in Table~\ref{tab-metrics}. A dash indicates that the source provides no corresponding value.}\label{tab-public}
\centering\footnotesize
\setlength{\tabcolsep}{4pt}
\begin{tabular*}{\textwidth}{@{\extracolsep{\fill}}llrrrrrr}
\toprule
&& \multicolumn{3}{c}{Glaucoma (REFUGE2)} & \multicolumn{3}{c}{Heart disease (MITEA)}\\
\cmidrule(lr){3-5}\cmidrule(lr){6-8}
Family & Method & bAcc & F1 & AUC & bAcc & F1 & AUC\\
\midrule
\multirow{6}{*}{\parbox{1.5cm}{General multimodal}}
 & GPT-4o \cite{gpt4o} &56.4&21.1&---&56.8&28.1&---\\
 & Janus-Pro-7B \cite{janus} &53.4&13.3&---&52.3&10.7&---\\
 & LLaVA-Med \cite{llavamed} &50.0&0.0&---&50.0&0.0&---\\
 & BioMedCLIP \cite{biomedclip} &58.1&21.3&---&47.0&37.8&---\\
 & Qwen2.5-VL-7B \cite{qwen25vl} &54.3&16.3&---&50.0&0.0&---\\
 & InternVL2.5-8B \cite{internvl25} &51.8&13.8&---&49.7&3.6&---\\
\cmidrule(lr){1-8}
\multirow{3}{*}{\parbox{1.5cm}{Medical agents}}
 & MedAgents \cite{medagents} &52.1&8.9&---&51.1&15.9&---\\
 & MMedAgent \cite{mmedagent} &52.4&16.3&---&55.0&26.7&---\\
 & MDAgents \cite{mdagents} &56.8&22.2&---&57.2&30.3&---\\
\cmidrule(lr){1-8}
\multirow{4}{*}{\parbox{1.5cm}{Task-specific}}
 & RetiZero \cite{retizero} &50.8&18.4&---&---&---&---\\
 & VisionUnite \cite{visionunite} &85.8&73.1&---&---&---&---\\
 & VUNO EYE TEAM \cite{refuge2} &---&---&88.3&---&---&---\\
 & MIG \cite{refuge2} &---&---&87.6&---&---&---\\
\cmidrule(lr){1-8}
\multirow{1}{*}{Tool agent} & MedAgent-Pro \cite{medagentpro} &90.4&76.4&95.1&77.8&72.3&---\\
\midrule
\multirow{4}{*}{\parbox{1.5cm}{\textbf{MedRSI}}}
 & Round 0 &\est{56.2}&\est{22.6}&\est{57.9}&\est{56.5}&\est{32.5}&\est{58.4}\\
 & Round 5 &\est{80.9}&\est{58.0}&\est{86.4}&\est{64.1}&\est{52.8}&\est{69.2}\\
 & Round 10 &\est{91.2}&\est{77.6}&\est{95.6}&\est{78.2}&\est{72.1}&\est{84.1}\\
 & Round 20 &\est{\textbf{94.1}}&\est{\textbf{83.2}}&\est{\textbf{97.2}}&\est{\textbf{82.3}}&\est{\textbf{77.3}}&\est{\textbf{88.6}}\\
\bottomrule
\end{tabular*}
\end{table*}

\begin{table*}[!t]
\caption{\textbf{Complete diagnostic metric panel across improvement rounds.} All values are percentages except the Matthews correlation coefficient (MCC) and the expected calibration error (ECE), which are reported on their native scales. Clinical cost is the mean consequence-weighted error per 100 test cases. Severe errors are errors graded serious or critical as a percentage of all test cases. Values are means over \est{five} independent trajectories, with 95\% confidence intervals from patient-level bootstrap resampling given for balanced accuracy. Predictive values and F1 depend on the test-partition prevalence, which is \est{11.5\%} for glaucoma and \est{36\%} for heart disease.}\label{tab-metrics}
\centering\footnotesize
\setlength{\tabcolsep}{3.4pt}
\begin{tabular*}{\textwidth}{@{\extracolsep{\fill}}llrrrrrrrrr}
\toprule
Task & Round & bAcc (95\% CI) & Sens. & Spec. & PPV & NPV & MCC & ECE & Cost & Severe\\
\midrule
\multirow{4}{*}{Glaucoma}
 & 0 &\est{56.2 (53.6--58.8)}&\est{28.4}&\est{84.0}&\est{18.7}&\est{90.0}&\est{0.10}&\est{0.29}&\est{12.4}&\est{9.8}\\
 & 5 &\est{80.9 (78.2--83.6)}&\est{71.6}&\est{90.2}&\est{48.7}&\est{96.1}&\est{0.53}&\est{0.17}&\est{5.9}&\est{4.4}\\
 & 10 &\est{91.2 (89.6--92.8)}&\est{87.3}&\est{95.1}&\est{69.8}&\est{98.3}&\est{0.75}&\est{0.09}&\est{3.0}&\est{2.1}\\
 & 20 &\est{\textbf{94.1} (92.8--95.4)}&\est{\textbf{92.0}}&\est{\textbf{96.2}}&\est{\textbf{75.9}}&\est{\textbf{98.9}}&\est{\textbf{0.81}}&\est{\textbf{0.05}}&\est{\textbf{2.5}}&\est{\textbf{1.5}}\\
\cmidrule(lr){1-11}
\multirow{4}{*}{\parbox{1.6cm}{Heart\\disease}}
 & 0 &\est{56.5 (53.1--59.9)}&\est{22.8}&\est{90.2}&\est{56.7}&\est{67.5}&\est{0.18}&\est{0.31}&\est{14.1}&\est{10.6}\\
 & 5 &\est{64.1 (60.5--67.7)}&\est{49.2}&\est{79.0}&\est{56.9}&\est{73.4}&\est{0.29}&\est{0.24}&\est{10.2}&\est{7.5}\\
 & 10 &\est{78.2 (75.4--81.0)}&\est{72.5}&\est{83.9}&\est{71.7}&\est{84.4}&\est{0.56}&\est{0.14}&\est{5.8}&\est{4.0}\\
 & 20 &\est{\textbf{82.3} (79.8--84.8)}&\est{\textbf{78.4}}&\est{\textbf{86.2}}&\est{\textbf{76.2}}&\est{\textbf{87.6}}&\est{\textbf{0.64}}&\est{\textbf{0.10}}&\est{\textbf{4.3}}&\est{\textbf{2.8}}\\
\bottomrule
\end{tabular*}
\end{table*}

\begin{figure*}[!t]
\centering
\includegraphics[width=\textwidth]{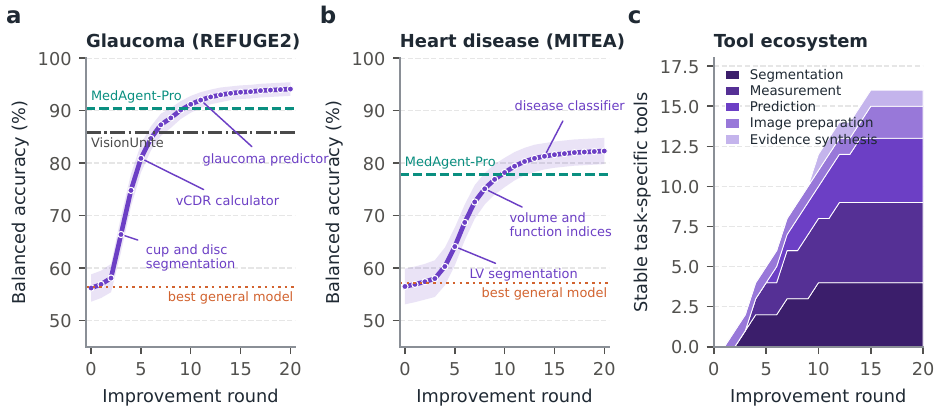}
\caption{\textbf{A general agent grows into a medical diagnostic system.} \textbf{a},\textbf{b}, Test balanced accuracy across improvement rounds for glaucoma (REFUGE2) and heart disease (MITEA). Lines show the mean over \est{five} trajectories and shading shows 95\% confidence intervals. Horizontal references reproduce MedAgent-Pro, the best general multimodal model and the task-specific VisionUnite model as reported by MedAgent-Pro\cite{medagentpro}. Annotations mark the registration round of representative tools. \textbf{c}, Cumulative number of stable task-specific tools across both tasks by capability category. General execution and retrieval utilities present in the initial agent are excluded from the count.}\label{fig-growth}
\end{figure*}

We can see a clear trajectory of autonomous capability growth across both clinical tasks. The initial agent performs at the level of a general multimodal model, reaching balanced accuracies of only \est{56.2\%} on glaucoma and \est{56.5\%} on heart disease (Table~\ref{tab-public}). Its early execution records reveal similarly generic behaviors, including qualitative descriptions of the optic nerve head without quantitative measurement, inconsistent judgments of cup size across visually similar eyes, and echocardiographic assessments that describe chamber appearance without estimating volume or function. As MedRSI repeatedly converts these failures into new tools, balanced accuracy rises steadily over 20 rounds to \est{94.1\%} for glaucoma and \est{82.3\%} for heart disease, with F1 scores reaching \est{83.2\%} and \est{77.3\%}, respectively (Fig.~\ref{fig-growth}a,b). The evolving agent surpasses the strongest general multimodal model after only \est{three} rounds on glaucoma and \est{four} rounds on heart disease, the task-specific VisionUnite model after \est{six} rounds, and MedAgent-Pro after \est{nine} and \est{ten} rounds. By the final round, it exceeds MedAgent-Pro by \est{3.7} and \est{4.5} balanced-accuracy points. This improvement is consistent across \est{all five} independent trajectories, with final balanced accuracies ranging from \est{93.4\%} to \est{94.7\%} for glaucoma and from \est{81.2\%} to \est{83.1\%} for heart disease.

The improvement is also not driven by a single favorable metric. As shown in Table~\ref{tab-metrics}, glaucoma sensitivity increases from \est{28.4\%} to \est{92.0\%} while specificity rises from \est{84.0\%} to \est{96.2\%}, with the Matthews correlation coefficient reaching \est{0.81}. Calibration improves in parallel, with the expected calibration error falling from \est{0.29} to \est{0.05} as the agent progressively replaces qualitative impressions with measured clinical indicators. Patient-level paired permutation tests confirm significant improvements from round 0 to round 20 on both tasks (\est{$P<0.001$}), as well as over the matched MedAgent-Pro reruns described in the Methods, which reach \est{89.6\%} and \est{77.1\%} balanced accuracy (\est{$P=0.004$} for glaucoma and \est{$P=0.011$} for heart disease). Taken together, these results show that MedRSI does more than optimize the performance of a fixed medical agent. Starting from a weak general-purpose agent, it progressively discovers and accumulates task-relevant medical capabilities through its own failures, eventually transforming itself into a substantially stronger specialist and surpassing manually engineered medical agents. The consistent gains across tasks, trajectories, error types, and calibration further demonstrate that this self-improvement is sustained and reproducible rather than the result of isolated tool discoveries or favorable evaluation noise.

The performance trajectory closely follows the acquisition of specific medical capabilities. On glaucoma, performance rises sharply after the agent registers optic cup and disc segmentation at round \est{4} and a vertical cup-to-disc ratio calculator at round \est{5}, and continues to improve as it acquires rim width measurement, peripapillary atrophy grading, and a glaucoma risk predictor that integrates quantitative measurements with image features. A similar progression emerges for heart disease, where gains follow the acquisition of left ventricular segmentation at round \est{5}, volume and ejection fraction estimation at round \est{7}, wall thickness and mass indices at round \est{9}, and a disease classifier at round \est{14}. Rather than expanding continuously, the stable toolkit grows in discrete steps as candidate capabilities complete their cohort trials (Fig.~\ref{fig-growth}c), ultimately reaching \est{16} task-specific tools by round 20, including \est{four} segmentation, \est{five} measurement, \est{four} prediction, \est{two} image preparation, and \est{one} evidence synthesis tool. Importantly, these retained tools represent only a small fraction of what the agent explores. Across 20 rounds, MedRSI proposes \est{118} candidate tools, of which \est{41} pass the discovery threshold and enter the experimental pool, while only \est{16} survive subsequent cohort trials and become persistent capabilities. This trajectory requires \est{212} accelerator hours and \est{1.9 billion} reasoning tokens per run, with \est{68\%} of the compute devoted to model training during candidate development. These results show that MedRSI improves not by simply accumulating an ever-growing collection of tools, but through a selective process of capability evolution in which many possibilities are explored, substantially fewer are experimentally validated, and only repeatedly beneficial capabilities become part of the persistent medical agent.

\subsection{Recursive self-improvement reconstructs an expert-designed tool ecosystem}

\begin{figure*}[!t]
\centering
\includegraphics[width=\textwidth]{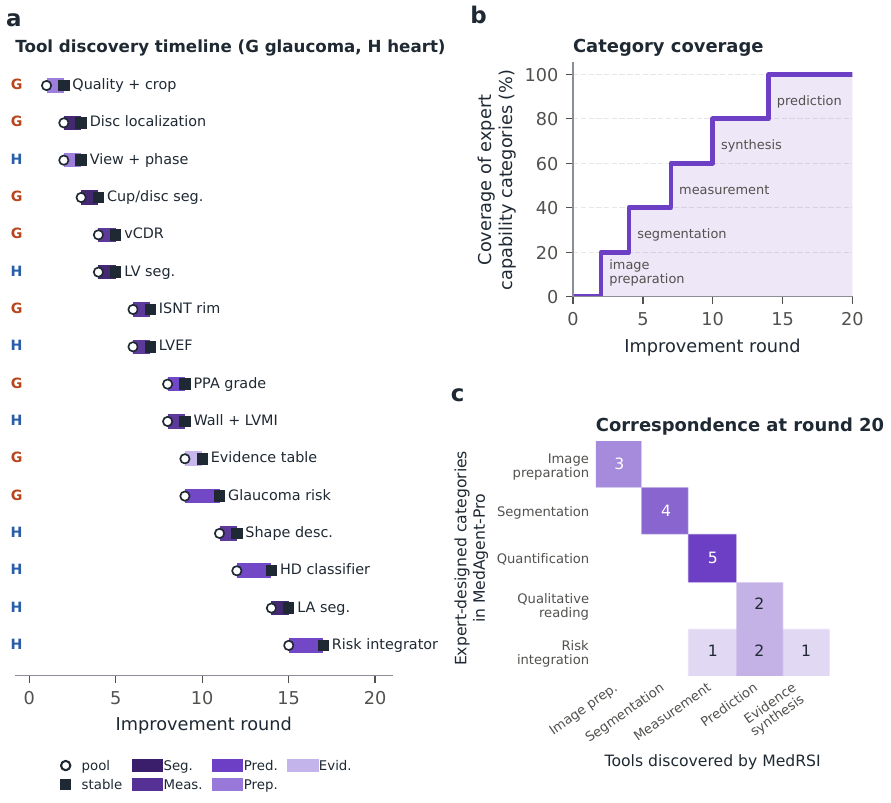}
\caption{\textbf{MedRSI rediscovers expert-designed clinical capabilities.} \textbf{a}, Timeline of the \est{16} stable tools acquired on the public tasks. Open circles mark entry into the experimental pool, filled squares mark registration into the stable agent and bar colour indicates the capability category. \textbf{b}, Coverage of the five expert-designed capability categories in MedAgent-Pro across improvement rounds. \textbf{c}, Correspondence between expert-designed categories and tools discovered by MedRSI at round 20. Cell values give the number of discovered tools that execute the corresponding expert function on held-out cases.}\label{fig-capabilities}
\end{figure*}

Remarkably, the tools invented by MedRSI without access to the MedAgent-Pro toolkit closely resemble those previously designed by human experts for the same clinical tasks. MedAgent-Pro, for example, uses a Medical SAM Adapter \cite{msa} for optic cup and disc segmentation, a coding agent to derive cup-to-disc ratio and rim thickness from segmentation masks, guideline-based assessment of the peripapillary region, and risk-weighted integration of these indicators \cite{medagentpro}. MedRSI independently rediscovers a closely related capability structure through self-improvement. As shown in Fig.~\ref{fig-capabilities}a, image preparation capabilities emerge first at rounds \est{1} and \est{2}, when the agent discovers that cropping the optic disc region and selecting an end-diastolic echocardiographic view improve downstream diagnosis. Segmentation capabilities emerge next at rounds \est{3}--\est{5}, followed by quantitative measurement at rounds \est{4}--\est{9}, and prediction and evidence synthesis from round \est{9} onward. This progression mirrors the natural dependency structure of clinical diagnosis, where quantitative measurements depend on anatomical localization and segmentation, while higher-level prediction depends in turn on the resulting measurements and imaging evidence. The emergence of this hierarchy without a predefined medical toolkit shows that MedRSI can autonomously recover clinically meaningful capability structures that previously required substantial human expertise to design.

We quantified this rediscovery through tool category coverage, defined as the fraction of expert-designed capability categories that at least one registered MedRSI tool executes on held-out cases. The category list was fixed before any MedRSI trajectory was reviewed and contains image preparation, segmentation, quantification, qualitative reading and risk integration. Coverage rises from \est{0\%} at round 0 to \est{20\%} at round \est{2}, \est{40\%} at round \est{4}, \est{60\%} at round \est{7}, \est{80\%} at round \est{10} and \est{100\%} at round \est{14} (Fig.~\ref{fig-capabilities}b). The correspondence at round 20 shows that each expert category is populated by between \est{two} and \est{five} discovered tools (Fig.~\ref{fig-capabilities}c). Without access to the manually designed MedAgent-Pro toolkit, MedRSI progressively rediscovered the same classes of capabilities that had previously been introduced by human experts.

The models autonomously trained by MedRSI achieve quality comparable to their expert-built counterparts. Starting from a generic U-Net template \cite{unet} and REFUGE2 training masks, the agent develops an optic cup and disc segmentation model that reaches Dice scores of \est{0.87} and \est{0.95}, respectively, on the test partition, with the resulting vertical cup-to-disc ratio measurements correlating strongly with reference values at \est{$r=0.91$}. Similarly, the autonomously developed left ventricular segmentation model achieves a Dice score of \est{0.90} on MITEA and supports ejection fraction estimation with a mean absolute error of \est{5.8} percentage points.
A representative development trajectory illustrates how such capabilities emerge directly from diagnostic failures. At round 3, MedRSI identifies \est{27} prioritized glaucoma failures in which the agent repeatedly describes the optic cup as ``moderately enlarged,'' despite reference cup-to-disc ratios ranging widely from \est{0.42} to \est{0.81}. Reflection traces these failures to the absence of quantitative cup measurement and identifies cup and disc segmentation as the missing capability. The tool builder then prepares the available training masks, instantiates and trains a segmentation network, evaluates it on the discovery cohort, and packages the resulting checkpoint as a callable tool. In the following round, MedRSI builds on this newly acquired capability by composing the cup and disc masks with diameter measurement to create a vertical cup-to-disc ratio tool that returns the quantitative ratio, anatomical contours, and a quality indicator. The development record preserves the motivating failures, model version, derived measurements, and cohort-level evaluation, providing a complete trace from observed clinical failure to a validated new capability. This example demonstrates how MedRSI can translate qualitative diagnostic weaknesses into concrete technical requirements and autonomously construct the specialized models and tools needed to address them.

The full inventory of registered tools is given in Table~\ref{tab-toolkit}, with the invention route, the round of registration, the resources each tool consumed and its intrinsic quality on held-out cases. Segmentation tools are evaluated with the Dice coefficient against the public annotations, measurement tools with the Pearson correlation against reference indicators, and prediction tools with balanced accuracy on the discovery cohort. Mean trial gain records the paired change in task balanced accuracy that the tool produced across its three trial cohorts, which is the quantity that governs registration. The inventory shows that the largest single gains come from the first segmentation and measurement tools on each task, and that later tools contribute smaller and steadier increments as the evidence available to the planner becomes more complete.

\begin{table*}[!t]
\caption{\textbf{Inventory of the stable tools acquired on the public tasks.} Route indicates code generation (C), tool composition (T) or model development (M). Round gives the improvement round at which the tool moved from the experimental pool into the stable registry. Intrinsic quality is the Dice coefficient for segmentation, the Pearson correlation against the reference indicator for measurement, and discovery-cohort balanced accuracy for prediction and synthesis. Mean trial gain is the paired change in task balanced accuracy across the three trial cohorts. All values are means over \est{five} trajectories.}\label{tab-toolkit}
\centering\footnotesize
\setlength{\tabcolsep}{2.8pt}
\begin{tabular*}{\textwidth}{@{\extracolsep{\fill}}llllrlrr}
\toprule
Tool & Task & Category & Route & Rd. & Resource & Quality & Gain\\
\midrule
Fundus quality and disc crop & G & Image prep. & C &\est{2}&\est{rule based}&\est{0.98}&\est{+0.9}\\
Optic disc localization & G & Segmentation & M &\est{3}&\est{600 images}&\est{0.94}&\est{+2.4}\\
Optic cup and disc segment. & G & Segmentation & M &\est{4}&\est{600 masks}&\est{0.87}&\est{+7.6}\\
Vertical cup-to-disc ratio & G & Measurement & T &\est{5}&\est{composition}&\est{0.91}&\est{+6.1}\\
Rim width by ISNT sector & G & Measurement & T &\est{7}&\est{composition}&\est{0.84}&\est{+2.2}\\
Peripapillary atrophy grading & G & Prediction & M &\est{9}&\est{600 images}&\est{0.81}&\est{+1.7}\\
Evidence table formatter & G & Synthesis & C &\est{10}&\est{rule based}&\est{---}&\est{+1.1}\\
Glaucoma risk predictor & G & Prediction & M &\est{11}&\est{600 cases}&\est{0.92}&\est{+2.8}\\
\cmidrule(lr){1-8}
Echo view and phase selection & H & Image prep. & C &\est{3}&\est{rule based}&\est{0.95}&\est{+1.2}\\
Left ventricular segmentation & H & Segmentation & M &\est{5}&\est{86 volumes}&\est{0.90}&\est{+5.8}\\
LV volume and ejection fraction & H & Measurement & T &\est{7}&\est{composition}&\est{0.88}&\est{+4.9}\\
Wall thickness and mass index & H & Measurement & T &\est{9}&\est{composition}&\est{0.82}&\est{+2.6}\\
Chamber shape descriptors & H & Measurement & C &\est{12}&\est{rule based}&\est{0.79}&\est{+1.4}\\
Heart disease classifier & H & Prediction & M &\est{14}&\est{86 subjects}&\est{0.83}&\est{+3.1}\\
Left atrial segmentation & H & Segmentation & M &\est{15}&\est{86 volumes}&\est{0.85}&\est{+1.3}\\
Risk integrator & H & Prediction & M &\est{17}&\est{86 cases}&\est{0.84}&\est{+1.6}\\
\bottomrule
\end{tabular*}
\end{table*}

\subsection{MedRSI invents new clinical AI models for previously unseen tasks}

The second experiment asks whether MedRSI can extend an already mature medical agent to clinical problems that its original designers never anticipated. We initialize $H_0$ as MedAgent-Pro with its complete expert-designed toolkit and expose it to two previously unseen clinical settings defined by private patient cohorts. The first is multimodal glaucoma diagnosis from linked fundus photography, optical coherence tomography, and visual field examinations collected across \est{two} tertiary eye centres. The cohort contains \est{1,898} eyes from \est{1,236} patients, with a glaucoma prevalence of \est{38\%}; \est{486} eyes form the test set and the \est{1,412} development eyes are split into training, discovery and trial partitions as described in the Methods. The second is left ventricular ejection fraction prediction from contrast-enhanced echocardiography, comprising \est{318} contrast studies and \est{2,940} noncontrast studies from \est{1,874} patients. Of these, \est{96} contrast studies form a held-out test set with cardiac magnetic resonance measurements as the reference standard, \est{42} paired studies are reserved for translation evaluation, and \est{120} contrast and \est{2,716} noncontrast studies are available for model training. Performance on the cardiac task is measured by mean absolute error normalized by the training-set standard deviation of ejection fraction, with lower values indicating more accurate prediction.

\begin{table*}[!t]
\caption{\textbf{Multimodal glaucoma diagnosis.} All values are percentages. Single-modality rows use the same reasoning agent restricted to one input. Late fusion combines the three single-modality probabilities with a fitted logistic model. The task-specific fusion model is trained by the study team on the same labelled training eyes with a development budget matched to the total budget consumed by MedRSI. Values are means over \est{five} runs with 95\% confidence intervals from patient-level bootstrap resampling.}\label{tab-glaucoma-mm}
\centering\footnotesize
\setlength{\tabcolsep}{4pt}
\begin{tabular*}{\textwidth}{@{\extracolsep{\fill}}llrrrrr}
\toprule
Family & Method & bAcc (95\% CI) & F1 & AUC & Sens. & Spec.\\
\midrule
General & GPT-4o &\est{68.4 (65.1--71.7)}&\est{60.9}&\est{72.6}&\est{61.3}&\est{75.5}\\
\cmidrule(lr){1-7}
\multirow{3}{*}{\parbox{1.4cm}{Single modality}}
 & Fundus only &\est{78.1 (75.2--81.0)}&\est{72.9}&\est{84.9}&\est{73.2}&\est{83.0}\\
 & OCT only &\est{80.6 (77.8--83.4)}&\est{75.9}&\est{87.2}&\est{76.9}&\est{84.3}\\
 & Visual field only &\est{75.3 (72.1--78.5)}&\est{69.4}&\est{81.7}&\est{70.1}&\est{80.5}\\
\cmidrule(lr){1-7}
\multirow{3}{*}{\parbox{1.5cm}{Multimodal}}
 & Late fusion &\est{83.9 (81.2--86.6)}&\est{80.0}&\est{90.4}&\est{80.6}&\est{87.2}\\
 & MedAgent-Pro &\est{79.2 (76.4--82.0)}&\est{74.2}&\est{85.8}&\est{74.5}&\est{83.9}\\
 & Task-specific fusion &\est{87.8 (85.3--90.3)}&\est{84.8}&\est{93.1}&\est{85.2}&\est{90.4}\\
\cmidrule(lr){1-7}
\multirow{3}{*}{\textbf{MedRSI}}
 & Round 0 &\est{79.2 (76.4--82.0)}&\est{74.2}&\est{85.8}&\est{74.5}&\est{83.9}\\
 & Round 6 &\est{88.4 (86.1--90.7)}&\est{85.6}&\est{94.0}&\est{86.1}&\est{90.7}\\
 & Round 12 &\est{\textbf{92.1} (90.2--94.0)}&\est{\textbf{90.2}}&\est{\textbf{96.4}}&\est{\textbf{90.3}}&\est{\textbf{93.9}}\\
\bottomrule
\end{tabular*}
\end{table*}

\begin{table*}[!t]
\caption{\textbf{Ejection fraction prediction from contrast echocardiography.} Normalized mean absolute error (nMAE) divides the absolute error by the training-set standard deviation of ejection fraction, and lower values indicate better prediction. MAE and bias are given in ejection fraction points against the cardiac magnetic resonance reference. The Pearson correlation and the intraclass correlation coefficient (ICC) assess agreement with the reference. Within 5 points is the percentage of test studies whose predicted ejection fraction falls inside the prespecified clinical tolerance. Values are means over \est{five} runs with 95\% confidence intervals from patient-level bootstrap resampling.}\label{tab-echo}
\centering\footnotesize
\setlength{\tabcolsep}{4pt}
\begin{tabular*}{\textwidth}{@{\extracolsep{\fill}}lrrrrrr}
\toprule
Method & nMAE (95\% CI) & MAE & Bias & $r$ & ICC & Within 5 pts\\
\midrule
GPT-4o &\est{1.00 (0.92--1.08)}&\est{9.3}&\est{$-4.1$}&\est{0.34}&\est{0.29}&\est{31.2}\\
MedAgent-Pro &\est{0.85 (0.78--0.92)}&\est{7.9}&\est{$-3.2$}&\est{0.52}&\est{0.48}&\est{42.7}\\
Noncontrast predictor applied directly &\est{0.92 (0.84--1.00)}&\est{8.6}&\est{$-3.8$}&\est{0.44}&\est{0.40}&\est{37.5}\\
Contrast-only predictor &\est{0.74 (0.67--0.81)}&\est{6.9}&\est{$-2.1$}&\est{0.66}&\est{0.63}&\est{51.0}\\
Synthetic-only predictor &\est{0.79 (0.72--0.86)}&\est{7.3}&\est{$-2.6$}&\est{0.61}&\est{0.58}&\est{47.9}\\
Task-specific regressor &\est{0.66 (0.60--0.72)}&\est{6.1}&\est{$-1.7$}&\est{0.74}&\est{0.71}&\est{58.3}\\
\midrule
\textbf{MedRSI}, round 0 &\est{0.85 (0.78--0.92)}&\est{7.9}&\est{$-3.2$}&\est{0.52}&\est{0.48}&\est{42.7}\\
\textbf{MedRSI}, round 8 &\est{0.63 (0.57--0.69)}&\est{5.9}&\est{$-1.4$}&\est{0.77}&\est{0.74}&\est{61.5}\\
\textbf{MedRSI}, round 12 &\est{\textbf{0.55} (0.50--0.60)}&\est{\textbf{5.1}}&\est{$\mathbf{-0.9}$}&\est{\textbf{0.83}}&\est{\textbf{0.81}}&\est{\textbf{68.8}}\\
\bottomrule
\end{tabular*}
\end{table*}

\begin{figure*}[!t]
\centering
\includegraphics[width=\textwidth]{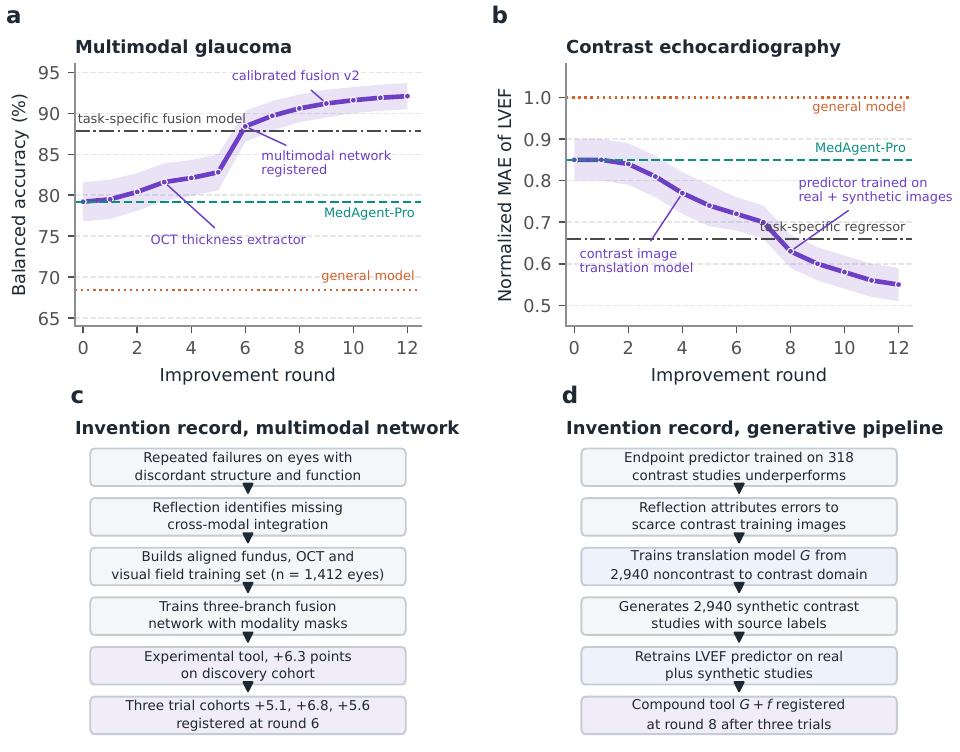}
\caption{\textbf{Invention of new clinical models on previously unseen tasks.} \textbf{a}, Test balanced accuracy for multimodal glaucoma diagnosis across improvement rounds, starting from MedAgent-Pro. \textbf{b}, Normalized mean absolute error of ejection fraction from contrast echocardiography across improvement rounds. Lines show means over \est{five} runs and shading shows 95\% confidence intervals. Horizontal references show MedAgent-Pro, the general multimodal model and the task-specific model trained on the same data. Annotations mark the registration of the tools that produce each change. \textbf{c}, Development record of the multimodal glaucoma network. \textbf{d}, Development record of the generative pipeline for contrast echocardiography, in which a translation model expands the contrast training set and a predictor is retrained on real and synthetic studies.}\label{fig-newtasks}
\end{figure*}

On multimodal glaucoma, MedAgent-Pro begins at \est{79.2\%} balanced accuracy (Table~\ref{tab-glaucoma-mm}). Its existing fundus tools remain effective, but errors concentrate in cases where structural and functional evidence disagree, such as a normal-appearing optic disc accompanied by an early arcuate field defect or a large physiological cup with a preserved visual field. Such discordant cases account for \est{71\%} of the prioritized failures during the first three rounds. MedRSI initially responds by expanding modality-specific capabilities, adding an optical coherence tomography thickness extractor and a visual field index parser during rounds \est{1}--\est{3}, which raises balanced accuracy to \est{81.6\%}. At round \est{4}, however, reflection identifies a deeper limitation: the modalities are available but still interpreted independently. MedRSI therefore proposes a dedicated multimodal diagnostic capability. The tool builder aligns fundus images, retinal nerve fibre layer thickness maps, and visual field total-deviation grids, and trains a three-branch fusion network with modality masks for missing inputs. The resulting candidate improves balanced accuracy by \est{6.3} points on the discovery cohort and subsequently by \est{5.1}, \est{6.8}, and \est{5.6} points across three independent trial cohorts. It is registered at round \est{6}, raising overall balanced accuracy to \est{88.4\%} (Fig.~\ref{fig-newtasks}a,c). A recalibrated second version is registered at round \est{9}, bringing the final agent to \est{92.1\%} balanced accuracy and \est{90.2\%} F1, compared with \est{87.8\%} balanced accuracy for a task-specific fusion model manually developed by the study team on the same data.

The final agent relies extensively on this autonomously acquired capability. The multimodal network is invoked in \est{94\%} of test cases and provides the highest-weighted evidence in \est{78\%}, while the remaining evidence is supplied by the inherited fundus tools and newly developed modality extractors. The single-modality controls in Table~\ref{tab-glaucoma-mm} further isolate the benefit of multimodal integration. Restricting the agent to a single modality yields balanced accuracies of \est{78.1\%} with fundus photography, \est{80.6\%} with optical coherence tomography, and \est{75.3\%} with visual fields, while fitted late fusion of their predictions reaches \est{83.9\%}. In contrast, the multimodal network developed by MedRSI reaches \est{92.1\%} balanced accuracy and an area under the curve of \est{96.4\%}. The advantage is particularly pronounced among discordant cases, where balanced accuracy reaches \est{86.7\%} compared with \est{71.4\%} for late fusion, while sensitivity for early glaucoma with a normal-appearing disc increases to \est{88.2\%} from \est{64.9\%} with the inherited fundus tools. These results show that MedRSI does not merely append tools for previously unseen modalities. By analyzing where its existing clinical workflow fails, it can identify the need for cross-modal reasoning and autonomously develop, validate, and incorporate a new capability that extends a mature medical agent beyond the problems anticipated by its original designers.

On contrast echocardiography, MedRSI develops a more complex chain of two interdependent capabilities (Fig.~\ref{fig-newtasks}b,d). MedAgent-Pro initially applies its existing noncontrast ventricular tools to contrast-enhanced studies and achieves a normalized error of \est{0.85}, as the inverted intensity pattern produced by ventricular opacification disrupts segmentation models trained on noncontrast images \cite{contrastguideline}. At round \est{2}, MedRSI responds by training an ejection fraction predictor directly on the \est{120} contrast training studies, reducing the error to \est{0.81} and advancing the model into the experimental pool. Reflection on the remaining failures then identifies a more fundamental bottleneck: only a small contrast-enhanced training set is available, while \est{2,716} noncontrast training studies carry the same type of reference labels but cannot be directly used because of the domain shift. Rather than treating this data limitation as fixed, MedRSI attempts to create the missing training resource itself. At round \est{4}, it trains an image translation model that maps noncontrast studies into the contrast-enhanced domain, verifies anatomical consistency between source and generated frames using its existing ventricular segmentation capability, and produces \est{2,716} synthetic contrast studies that inherit the labels of their source examinations. At round \est{6}, the agent retrains its ejection fraction predictor jointly on real and synthetic contrast studies. The resulting compound capability improves normalized error by \est{0.09} on the discovery cohort and by \est{0.07}, \est{0.10}, and \est{0.08} across three subsequent trial cohorts, leading to registration at round \est{8}.

The final agent reaches a normalized error of \est{0.55}, corresponding to a mean absolute error of \est{5.1} ejection fraction points, compared with \est{0.66} for a task-specific regressor trained only on real contrast studies. Training on synthetic studies alone yields \est{0.79}, indicating that the benefit arises from augmenting rather than replacing the limited real data. The generated images achieve a structural similarity of \est{0.81} against paired real contrast frames in the \est{42} patients for whom both study types are available. Agreement with the cardiac magnetic resonance reference also improves consistently across the full metric panel in Table~\ref{tab-echo}: mean absolute error falls to \est{5.1} ejection fraction points with a bias of \est{$-0.9$} points, Pearson correlation reaches \est{0.83}, and the intraclass correlation coefficient reaches \est{0.81}. The proportion of studies within the prespecified clinical tolerance of \est{5} ejection fraction points increases from \est{42.7\%} to \est{68.8\%}. By comparison, directly applying the original noncontrast predictor to contrast studies produces a normalized error of \est{0.92}, confirming the domain shift identified during self-improvement. This case demonstrates a deeper form of recursive capability development: MedRSI identifies that its remaining limitation is not simply a missing predictor but a shortage of appropriate training data, autonomously creates a generative capability to overcome that bottleneck, and then uses the newly created resource to develop a stronger downstream clinical model.

\subsection{Clinical-cost prioritization directs self-improvement toward consequential failures}

\begin{figure*}[!t]
\centering
\includegraphics[width=\textwidth]{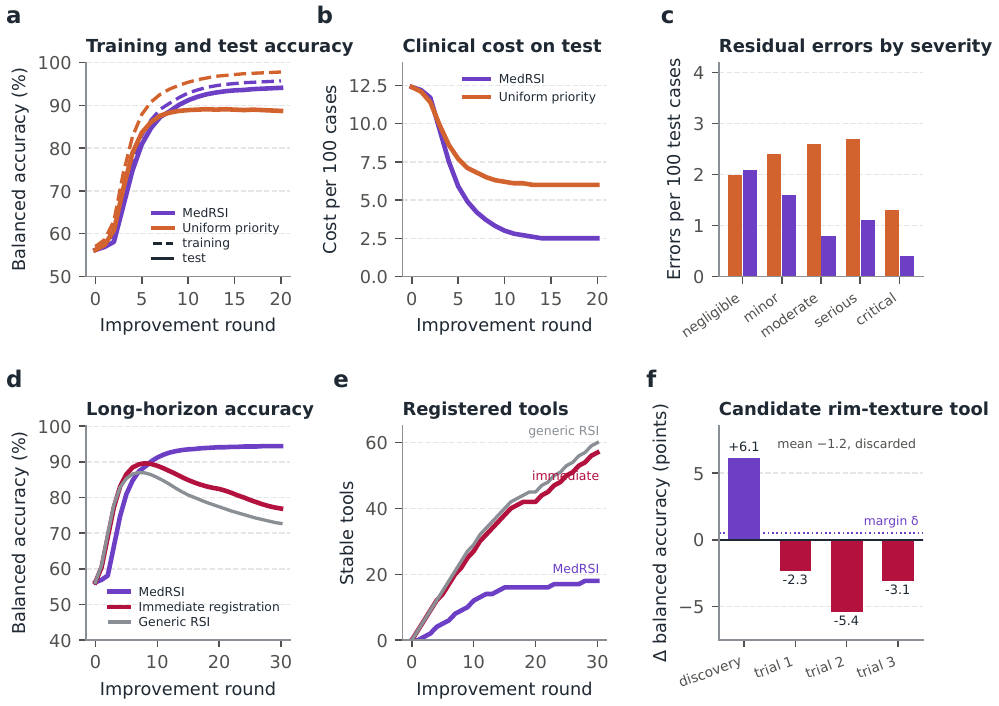}
\caption{\textbf{Clinical-cost prioritization and slow registration are required for medical recursive self-improvement.} \textbf{a}, Training and test balanced accuracy for the complete agent and the uniform priority ablation on the glaucoma trajectory. \textbf{b}, Mean clinical cost per 100 test cases. \textbf{c}, Residual test errors at round 20 by consequence grade. \textbf{d}, Test balanced accuracy over 30 rounds for the complete agent, immediate registration and generic recursive self-improvement, which combines uniform priority with immediate registration. \textbf{e}, Number of stable tools over the same horizon. \textbf{f}, Paired balanced accuracy change of one candidate tool across its discovery cohort and three trial cohorts. The dotted line marks the registration margin $\delta$.}\label{fig-ablation}
\end{figure*}

The third experiment isolates the contribution of the two clinical mechanisms through matched ablations on the glaucoma trajectory. All configurations use the same initial agent, patient partitions, candidate-development budget, and random seeds. We first remove clinical-cost-aware prioritization by setting the clinical weighting coefficient to zero, such that every diagnostic error receives equal priority for self-improvement. Under this setting, training performance initially improves faster than with the complete MedRSI framework and reaches \est{97.8\%} balanced accuracy by round 20. This apparent advantage, however, does not transfer to unseen patients. Test performance plateaus at \est{88.7\%} from round \est{11} onward (Fig.~\ref{fig-ablation}a), whereas the complete agent reaches \est{95.7\%} on training and \est{94.1\%} on test by round 20. The resulting train--test gap is only \est{1.6} points for MedRSI, compared with \est{9.1} points under uniform prioritization. The trajectory shows that treating every failure equally encourages RSI to optimize for the errors it encounters most frequently, producing tools that rapidly eliminate common training mistakes but contribute less to the rarer, clinically consequential failures that remain important across patients. Clinical-cost-aware prioritization instead changes what the agent chooses to improve, directing its finite self-improvement budget toward failures whose resolution is both clinically meaningful and more transferable beyond the cases that originally triggered the update.

\begin{table*}[!t]
\caption{\textbf{Ablation of the improvement mechanisms on the glaucoma trajectory.} Test balanced accuracy, F1 and AUC are measured at round 20, and long-horizon balanced accuracy at round 30. Clinical cost is the mean consequence-weighted error per 100 test cases. Severe errors are errors graded serious or critical as a percentage of all test cases. Stable tools are counted at round 20. Discard rate is the percentage of experimental tools removed before registration. All configurations share the initial agent, patient partitions, candidate development budget and random seeds. Values are means over \est{five} runs.}\label{tab-ablation}
\centering\footnotesize
\setlength{\tabcolsep}{3.2pt}
\begin{tabular*}{\textwidth}{@{\extracolsep{\fill}}lrrrrrrrr}
\toprule
Configuration & bAcc & F1 & AUC & Cost & Severe & Tools & Discard & Round 30\\
\midrule
Generic RSI &\est{77.4}&\est{61.8}&\est{83.1}&\est{9.2}&\est{6.4}&\est{45}&\est{0}&\est{72.7}\\
\cmidrule(lr){1-9}
Uniform failure priority &\est{88.7}&\est{74.6}&\est{92.8}&\est{6.0}&\est{4.0}&\est{15}&\est{61}&\est{88.3}\\
Immediate registration &\est{82.4}&\est{67.2}&\est{87.4}&\est{7.6}&\est{5.1}&\est{42}&\est{0}&\est{76.9}\\
Single trial cohort &\est{90.6}&\est{77.9}&\est{94.1}&\est{4.7}&\est{3.2}&\est{31}&\est{24}&\est{84.7}\\
\cmidrule(lr){1-9}
Without failure clustering &\est{90.2}&\est{77.1}&\est{93.6}&\est{3.9}&\est{2.6}&\est{14}&\est{66}&\est{90.1}\\
Code generation only &\est{79.8}&\est{63.4}&\est{85.2}&\est{7.1}&\est{5.3}&\est{9}&\est{58}&\est{79.4}\\
Without model development &\est{84.3}&\est{69.8}&\est{89.0}&\est{5.4}&\est{3.9}&\est{11}&\est{60}&\est{84.0}\\
Without tool composition &\est{89.1}&\est{75.3}&\est{93.0}&\est{4.1}&\est{2.9}&\est{13}&\est{63}&\est{89.0}\\
\cmidrule(lr){1-9}
\textbf{MedRSI} &\est{\textbf{94.1}}&\est{\textbf{83.2}}&\est{\textbf{97.2}}&\est{\textbf{2.5}}&\est{\textbf{1.5}}&\est{16}&\est{61}&\est{\textbf{94.4}}\\
\bottomrule
\end{tabular*}
\end{table*}

Clinical consequence weighting changes not only which failures MedRSI prioritizes, but also which capabilities it ultimately develops. Under uniform prioritization, \est{63\%} of the reflection budget during the first ten rounds is spent on borderline eyes labeled as glaucoma suspect, where diagnostic disagreement carries negligible or minor clinical consequence. In contrast, the complete agent devotes \est{58\%} of the same budget to high-consequence failures, particularly advanced glaucoma misclassified as normal and early glaucoma cases that would otherwise have been discharged. This difference directly shapes the emerging toolkit. Motivated by missed advanced cases, the complete agent develops a sector-specific rim width measurement tool at round \est{7} and a peripapillary atrophy grader at round \est{9}. Under uniform prioritization, the agent instead develops a second cup-to-disc threshold calibrator and a disc-size normalizer, further refining decisions around borderline cases. By round 20, this difference translates into a mean clinical cost of \est{2.5} per 100 test cases for the complete agent compared with \est{6.0} under uniform prioritization (Fig.~\ref{fig-ablation}b). Serious and critical errors account for only \est{1.5\%} of test cases, compared with \est{4.0\%} under uniform prioritization (Fig.~\ref{fig-ablation}c and Table~\ref{tab-ablation}). Missed advanced glaucoma, the most consequential error category, falls to \est{0.4\%} from \est{1.3\%}, while sensitivity for advanced disease increases from \est{92.6\%} to \est{97.9\%}. These results show that clinical-cost-aware prioritization changes the direction of self-improvement itself, steering MedRSI away from merely correcting the most common errors and toward acquiring capabilities that reduce clinically consequential failures.

The remaining ablations in Table~\ref{tab-ablation} examine how MedRSI turns these identified capability gaps into new tools. Restricting invention to code generation yields only \est{79.8\%} balanced accuracy, as the agent can manipulate and quantify existing evidence but cannot acquire fundamentally new perceptual capabilities. Removing model development while retaining tool composition reaches \est{84.3\%}, whereas retaining model development but removing composition reaches \est{89.1\%}. The three invention routes are therefore complementary, with model development accounting for \est{9.8} of the \est{14.3} balanced-accuracy points separating the code-only configuration from the complete agent. Failure clustering is similarly important for efficient capability discovery. When clustering is removed and the builder instead receives individual high-priority failures, performance reaches \est{90.2\%} with \est{14} registered tools, while the development records contain \est{four} near-duplicate candidates targeting the same underlying capability gap. Grouping related failures therefore helps MedRSI reason at the level of missing capabilities rather than individual mistakes, reducing redundant invention and producing a more effective persistent toolkit.

\subsection{Slow registration prevents tool accumulation from destabilizing long-horizon self-improvement}

The second ablation removes slow registration, admitting each candidate into the stable agent as soon as it improves the discovery endpoint. We extend all trajectories to 30 rounds to expose their long-horizon behavior. Immediate registration initially appears advantageous. It reaches \est{89.6\%} balanced accuracy by round \est{8}, compared with \est{88.6\%} for the complete agent, while accumulating \est{22} stable tools compared with only \est{9}. This early advantage, however, quickly reverses. As the toolkit continues to expand, balanced accuracy steadily declines to \est{82.4\%} at round \est{20} and \est{76.9\%} at round \est{30}, by which point the agent has accumulated \est{57} tools (Fig.~\ref{fig-ablation}d,e). The degradation is even stronger under generic recursive self-improvement, which combines uniform failure priority with immediate registration. Performance peaks at \est{87.0\%} at round \est{7} before falling to \est{72.7\%} by round \est{30}, despite the toolkit growing to \est{60} tools. In contrast, the complete MedRSI agent selectively registers only \est{18} tools over the same horizon and maintains \est{94.4\%} balanced accuracy at round \est{30}, with no round-to-round decline exceeding \est{0.3} points after round \est{12}. More capability accumulation therefore does not necessarily produce a stronger self-improving agent. Without controlling which discoveries become persistent, continued self-improvement can progressively degrade the system it is intended to improve.

The immediate-registration trajectory reveals how this degradation becomes self-reinforcing. One candidate tool that scores neuroretinal rim texture improves its discovery cohort by \est{6.1} points and is therefore registered immediately. Yet its contributions on the next three cohorts are \est{$-2.3$}, \est{$-5.4$}, and \est{$-3.1$} points (Fig.~\ref{fig-ablation}f), as the learned texture features are sensitive to camera model and illumination conditions specific to the discovery cohort. Once registered, however, the planner invokes this tool in \est{81\%} of subsequent glaucoma cases, and its outputs become part of the execution records consumed by later rounds of reflection. Within only \est{four} rounds, the agent develops \est{three} additional tools that post-process the flawed texture score, while an increasing fraction of prioritized failure clusters originate from cases in which that score was itself misleading. A locally beneficial but unstable discovery thus becomes part of the environment from which subsequent self-improvement learns, allowing one error to propagate into later trajectories and generate further misguided capabilities. This creates a destructive feedback loop in which poor tools produce poor trajectories, poor trajectories motivate further poor tool discoveries, and accumulated errors eventually drive the agent toward collapse.

Slow registration breaks this feedback loop before an unstable capability can become part of the persistent agent. Under the complete MedRSI framework, the same texture candidate enters the experimental pool but achieves a mean trial gain of \est{$-1.2$} points against a required margin of \est{0.5} and is discarded before any downstream capability can depend on it. Across the full 30-round trajectory, MedRSI rejects \est{43} experimental tools that would have satisfied the immediate-registration criterion based on their discovery cohorts alone, of which \est{29} subsequently exhibit negative mean gains during cohort trials. These results show that long-horizon RSI requires more than discovering improvements quickly. It requires controlling which improvements are allowed to become part of the system's future self-improvement process. By separating rapid capability discovery from conservative registration, MedRSI prevents transient gains from being recursively amplified into persistent failures and enables stable self-improvement over long horizons.

\subsection{Self-improvement is stable across settings, reasoning models and patient subgroups}

\begin{figure*}[!t]
\centering
\includegraphics[width=\textwidth]{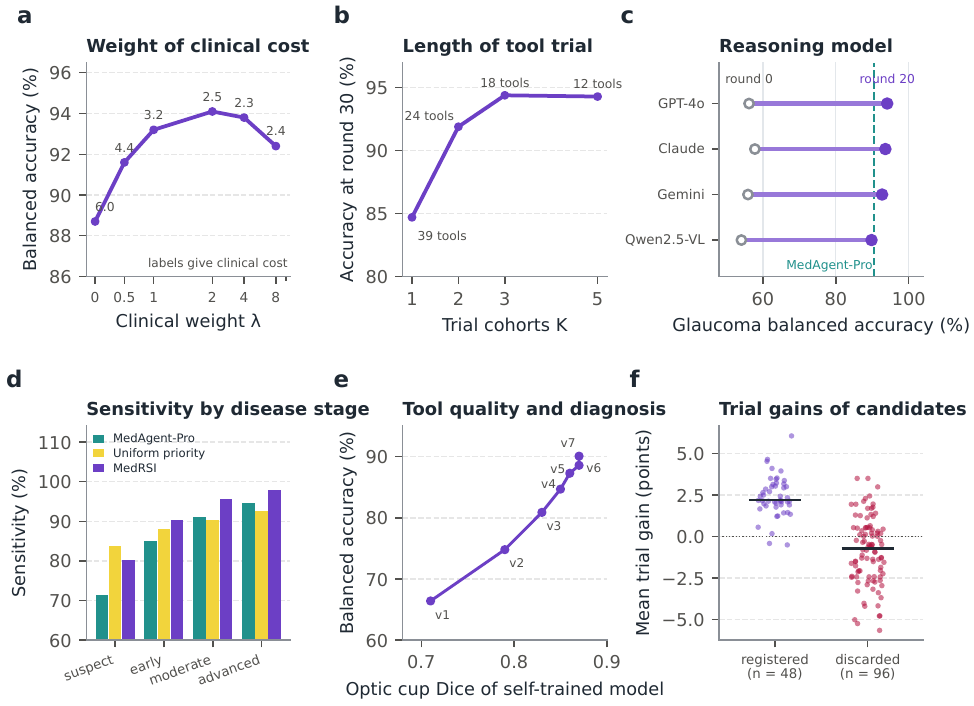}
\caption{\textbf{Sensitivity of MedRSI to its settings, its reasoning model and the patient population.} \textbf{a}, Balanced accuracy at round 20 across values of the clinical weight $\lambda$, with the mean clinical cost printed above each point. \textbf{b}, Balanced accuracy at round 30 across the number of trial cohorts $K$, with the number of registered tools printed above each point. \textbf{c}, Glaucoma balanced accuracy at rounds 0 and 20 for four reasoning models, with the MedAgent-Pro reference. \textbf{d}, Sensitivity by glaucoma stage for MedAgent-Pro, the uniform priority ablation and the complete agent. \textbf{e}, Agent balanced accuracy against the optic cup Dice coefficient of successive versions of the self-trained segmentation model. \textbf{f}, Mean trial gain of every candidate that reached the experimental pool on the public tasks, separated by registration outcome, with horizontal lines at the medians.}\label{fig-robust}
\end{figure*}

MedRSI remains robust across its two main protocol settings, as shown in Fig.~\ref{fig-robust}. Increasing the clinical weighting coefficient $\lambda$ raises round-20 balanced accuracy from \est{88.7\%} at $\lambda=0$ to a maximum of \est{94.1\%} at $\lambda=2$. Performance remains stable over a relatively broad range before declining to \est{92.4\%} at $\lambda=8$, where self-improvement becomes concentrated on a small number of severe cases at the expense of the broader error distribution. Clinical cost decreases monotonically as $\lambda$ increases and largely saturates beyond $\lambda=2$, indicating a broad operating region between \est{1} and \est{4} rather than dependence on a narrowly tuned value. The number of trial cohorts produces a similarly interpretable trade-off between the speed and stability of capability accumulation. Using a single trial cohort allows \est{39} tools to enter the persistent agent but yields only \est{84.7\%} balanced accuracy at round 30. Requiring three cohorts reduces the toolkit to \est{18} tools while maintaining \est{94.4\%}, whereas five cohorts further reduces it to \est{12} tools with a nearly identical \est{94.3\%}, at the cost of \est{two} additional rounds before each candidate can qualify (Fig.~\ref{fig-robust}a,b). Together, these results show that MedRSI is not sensitive to a single protocol configuration, while also revealing the expected trade-offs between clinical focus, discovery speed, and long-term stability.

The self-improvement process also transfers across reasoning backbones. Replacing the backbone with \est{Claude Sonnet 4.5} \cite{claudesonnet45}, \est{Gemini 2.5 Pro} \cite{gemini25}, and \est{Qwen2.5-VL-72B} \cite{qwen25vl} produces round-0 balanced accuracies ranging from \est{54.1\%} to \est{57.8\%} and round-20 accuracies from \est{89.8\%} to \est{93.6\%}, with \est{three of the four} evaluated backbones ultimately exceeding MedAgent-Pro (Fig.~\ref{fig-robust}c). More importantly, different reasoning models independently converge toward similar capability structures. Of the \est{16} tool functions identified across the trajectories, \est{11} are recovered by all four backbones and \est{14} by at least three. Segmentation and quantitative measurement capabilities emerge consistently across every backbone, while most variation appears only in the higher-level synthesis tools developed during later rounds. This convergence suggests that the capabilities discovered by MedRSI are not artifacts of a particular reasoning model, but reflect recurring capability gaps imposed by the underlying clinical tasks.

Subgroup analysis shows where the acquired capabilities act (Fig.~\ref{fig-robust}d). Sensitivity for advanced glaucoma reaches \est{97.9\%} for the complete agent, \est{92.6\%} under uniform priority and \est{94.6\%} for MedAgent-Pro, and sensitivity for moderate disease reaches \est{95.7\%}, \est{90.4\%} and \est{91.2\%}. The uniform priority configuration performs better on the glaucoma suspect subgroup, at \est{83.7\%} against \est{80.2\%}, which follows from the improvement budget it spends there. Performance is stable across acquisition sites, with balanced accuracy of \est{94.4\%} and \est{93.6\%} at the two glaucoma centres, and across image quality strata, with \est{95.1\%} for gradable and \est{89.8\%} for borderline-quality photographs. The quality of an individual tool translates into diagnostic benefit in a graded way. Successive versions of the self-trained optic cup model span Dice scores from \est{0.71} to \est{0.87}, and agent balanced accuracy rises from \est{66.4\%} to \est{90.1\%} across those versions (Fig.~\ref{fig-robust}e). Across the \est{205} candidates that reached the experimental pool on the public tasks over the five trajectories (\est{41} per trajectory, Table~\ref{tab-budget}), registered tools have a median mean trial gain of \est{2.2} points and discarded tools a median of \est{$-0.9$} points, with an overlap region between \est{0} and \est{1.5} points that the registration margin resolves (Fig.~\ref{fig-robust}f).

\begin{table*}[!t]
\caption{\textbf{Discovery activity and computational budget.} Candidates counts the development attempts the builder completed, pool counts those that passed the discovery threshold, and registered counts those that survived their trials. Accelerator hours are measured on \est{NVIDIA A100 80 GB} devices and cover candidate model training and evaluation. Reasoning tokens cover diagnosis, reflection, tool building and judging. Values are means over \est{five} runs for the public tasks and over \est{five} runs for each additional task.}\label{tab-budget}
\centering\footnotesize
\setlength{\tabcolsep}{3.2pt}
\begin{tabular*}{\textwidth}{@{\extracolsep{\fill}}lrrrrrrrr}
\toprule
Task & Rd. & Cand. & Pool & Reg. & GPU h & Tok. (B) & Days & h/tool\\
\midrule
Glaucoma (REFUGE2) &20&\est{64}&\est{22}&\est{8}&\est{96}&\est{0.9}&\est{2.1}&\est{12.0}\\
Heart disease (MITEA) &20&\est{54}&\est{19}&\est{8}&\est{116}&\est{1.0}&\est{2.4}&\est{14.5}\\
Multimodal glaucoma &12&\est{41}&\est{14}&\est{6}&\est{174}&\est{0.7}&\est{3.2}&\est{29.0}\\
Contrast echo &12&\est{38}&\est{13}&\est{5}&\est{265}&\est{0.6}&\est{4.6}&\est{53.0}\\
\midrule
Total &---&\est{197}&\est{68}&\est{27}&\est{651}&\est{3.2}&\est{12.3}&\est{24.1}\\
\bottomrule
\end{tabular*}
\end{table*}

The cost of the process is recorded in Table~\ref{tab-budget}. A complete public-task trajectory consumes \est{212} accelerator hours and \est{1.9 billion} reasoning tokens and completes in \est{4.5} days on a single node. The two additional clinical tasks are more expensive per registered tool, at \est{29.0} and \est{53.0} accelerator hours, because their inventions involve training multimodal and generative networks. Model development accounts for \est{68\%} of the accelerator hours and \est{21\%} of the tokens, and diagnosis of clinical experience batches accounts for \est{54\%} of the tokens. Across all four tasks the agent completed \est{197} development attempts, of which \est{68} entered the experimental pool and \est{27} were registered.

\section{Discussion}\label{sec-discussion}

A medical agent need not remain limited to a toolkit designed before deployment. Through recursive interaction with diagnostic experience, MedRSI progressively constructs its own specialized capabilities. Each improvement is embodied in an executable artifact with a training record, a calling specification and a trial history that a clinician or engineer can inspect, and each artifact enlarges the evidence available to the agent for subsequent patients. Segmentation supports measurement, measurement supports prediction and prediction supports the integration of multiple indicators into a diagnosis.

When initialized from a public general-purpose agent platform without the expert-designed toolkit used by MedAgent-Pro, the agent independently recovered the same capability classes, in an order that follows the dependency structure of the clinical task, and its final performance exceeded the manually engineered agent on both public benchmarks. Recursive self-improvement extended beyond rediscovery. On previously unseen clinical tasks the agent created new task-specific AI models, including a multimodal diagnostic network that integrates structure and function in glaucoma and a generative model that overcame scarce contrast echocardiography training data before training a downstream predictor. Both inventions arose from the agent's own analysis of where its errors concentrated and what resource was missing.

Our experiments also identify two properties required for this process in medicine. Errors must be weighted by clinical consequence, since improvement guided by error frequency alone learns the training distribution of mistakes and leaves the rare, consequential failures in place. Capability discovery must occur faster than capability consolidation, since a permanent toolkit that absorbs every promising candidate accumulates unstable tools whose outputs contaminate later reasoning and later discovery. Clinical-cost-aware failure prioritization and fast discovery with slow registration implement these properties with a single weighting coefficient and a single trial rule. The effectiveness of both mechanisms depends on the quality of reference diagnoses, the consistency of consequence scoring and the availability of independent evaluation cohorts, and prospective clinical studies will be needed to examine clinician use, workflow effects and patient outcomes of agents that continue to develop after deployment. The framework offers a route by which medical agents progress from using human-designed tools to building and consolidating their own clinical capabilities through experience.

\bibliography{medrsi}
\end{document}